\documentclass[journal]{IEEEtran}
\usepackage{amsmath,amsfonts,bm}

\def\eqref#1{equation~\ref{#1}}

\def\1{\bm{1}}

\def\vf{{\bm{f}}}

\DeclareMathAlphabet{\mathsfit}{\encodingdefault}{\sfdefault}{m}{sl}
\SetMathAlphabet{\mathsfit}{bold}{\encodingdefault}{\sfdefault}{bx}{n}

\def\sB{{\mathbb{B}}}

\def\sX{{\mathbb{X}}}

\usepackage[utf8]{inputenc} 
\usepackage[T1]{fontenc}    
\usepackage{url}            
\usepackage{booktabs}       
\usepackage{amsfonts}       
\usepackage{nicefrac}       
\usepackage{microtype}      

\usepackage{graphicx}
\usepackage{floatrow}
\usepackage{comment}
\usepackage{amsmath,amssymb} 
\usepackage{color}
\newfloatcommand{figurebox}{figure}[\nocapbeside][\dimexpr(\textwidth-\columnsep)/2\relax]
\newfloatcommand{tablebox}{table}[\nocapbeside][\dimexpr(\textwidth-\columnsep)/2\relax]
\usepackage{tabularx}
\usepackage{arydshln}

\usepackage{array}
\newcolumntype{C}[1]{>{\centering\arraybackslash}m{#1}}
\newcolumntype{R}[1]{>{\raggedleft\arraybackslash}m{#1}}
\newcolumntype{P}[1]{>{\raggedright\arraybackslash}p{#1}}
\newcolumntype{M}[1]{>{\centering\arraybackslash}m{#1}}
\usepackage{multirow}
\usepackage{algorithm}
\usepackage{algorithmic}

\usepackage{hyperref}
\usepackage{cleveref}
\floatstyle{ruled}
\restylefloat{algorithm}
\usepackage[font=footnotesize,labelfont=bf,labelsep=period]{caption}

\newcommand{\ie}{\textit{i}.\textit{e}., }
\newcommand{\eg}{\textit{e}.\textit{g}., }

\ifCLASSINFOpdf
\else
\fi
\begin{document}
%
\title{Structured Domain Adaptation with Online Relation Regularization for Unsupervised Person Re-ID}
%
%
%

\author{Yixiao~Ge,~Feng~Zhu,~Dapeng~Chen,~Rui~Zhao, \textit{Member, IEEE},\\Xiaogang~Wang, \textit{Member, IEEE},~and~Hongsheng~Li, \textit{Member, IEEE}%
\thanks{
Hongsheng Li is the corresponding author.
email: lihongsheng@gmail.com
}
\thanks{Yixiao Ge is with Multimedia Laboratory, The Chinese University of Hong Kong and ARC Lab, Tencent PCG.}
\thanks{
Feng Zhu, Dapeng Chen, and Rui Zhao are with Multimedia Laboratory, The Chinese University of Hong Kong.}
\thanks{
Xiaogang Wang and Hongsheng Li are with Multimedia Laboratory, The Chinese University of Hong Kong and Centre for Perceptual and Interactive Intelligence Limited, Hong Kong SAR, China.}
\thanks{
This work is supported in part by Centre for Perceptual and Interactive Intelligence Limited, in part by the General Research Fund through the Research Grants Council of Hong Kong under Grants (Nos. 14204021, 14207319, 14203118, 14208619), in part by Research Impact Fund Grant No. R5001-18, in part by CUHK Strategic Fund.
}
}

\markboth{IEEE TRANSACTIONS ON NEURAL NETWORKS AND LEARNING SYSTEMS}%
{Ge \MakeLowercase{\textit{et al.}}: Structured Domain Adaptation with Online Relation Regularization for Unsupervised Person Re-ID}
%



\maketitle

\begin{figure*}[!t]

\begin{center}
    \centering
    \footnotesize
     \begin{tabular}{C{8.5cm}C{8.5cm}}
     \begin{tabular}{c}
     \shortstack[l]{\textbf{(a)} Existing domain translation-based UDA methods  \cite{deng2018image,wei2018person,deng2018similarity,chen2019instance,tang2020cgan} \\ with ID-based regularizations.} \vspace{5pt} \\
     \includegraphics[width=0.9\linewidth]{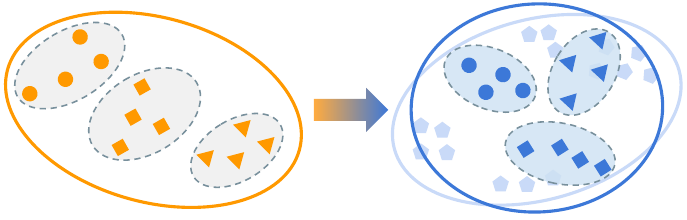}
     \end{tabular} &
     \begin{tabular}{c}
     \shortstack[l]{\textbf{(b)} Our proposed structured domain translation with a relation-\\consistency  regularization.}  \vspace{5pt} \\
     \includegraphics[width=0.9\linewidth]{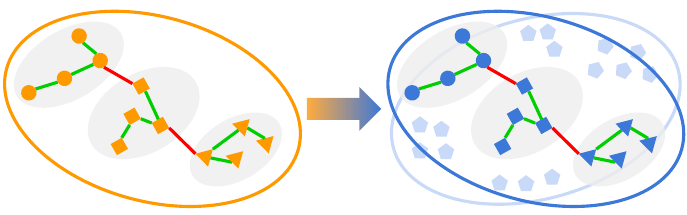}
     \end{tabular}
    \end{tabular}

    \begin{tabular}{c@{}}
   \includegraphics[width=0.95\linewidth]{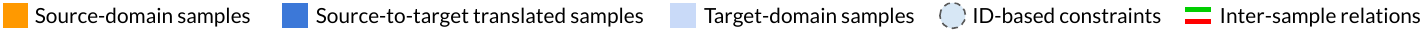} \\
    \end{tabular}   \vspace{8pt} \\

   \begin{tabular}{c@{}}
   \textbf{(c)} Domain-translated triplets by CycleGAN \cite{zhu2017unpaired}, SPGAN \cite{wei2018person} and our structured domain-translation method. \\
    \end{tabular}   \vspace{5pt}  \\
    \begin{tabular}{c@{\hspace{0.5mm}}c@{\hspace{0.5mm}}c@{\hspace{1.5mm}}c@{\hspace{0.5mm}}c@{\hspace{0.5mm}}c@{\hspace{1.5mm}}c@{\hspace{0.5mm}}c@{\hspace{0.5mm}}c@{\hspace{1.5mm}}c@{\hspace{0.5mm}}c@{\hspace{0.5mm}}c|c@{\hspace{0.5mm}}c@{\hspace{0.5mm}}c@{\hspace{1.5mm}}c@{\hspace{0.5mm}}c@{\hspace{0.5mm}}c@{\hspace{1.5mm}}c@{\hspace{0.5mm}}c@{\hspace{0.5mm}}c@{\hspace{1.5mm}}c@{\hspace{0.5mm}}c@{\hspace{0.5mm}}c}
   	\multicolumn{3}{c}{{\scriptsize Two IDs}} & \multicolumn{3}{c}{{\scriptsize Look similar}} & \multicolumn{3}{c}{{\scriptsize Look similar}} & \multicolumn{3}{c}{{\scriptsize Keep difference}} & \multicolumn{3}{c}{{\scriptsize Two IDs}} & \multicolumn{3}{c}{{\scriptsize Look different}} & \multicolumn{3}{c}{{\scriptsize Look different}} & \multicolumn{3}{c}{{\scriptsize Keep similarity}} \\
        \includegraphics[scale=0.14]{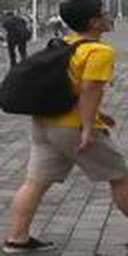} &
        \includegraphics[scale=0.14]{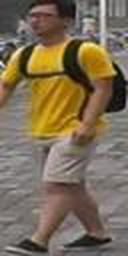} &
        \includegraphics[scale=0.14]{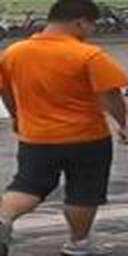} &
        \includegraphics[scale=0.14]{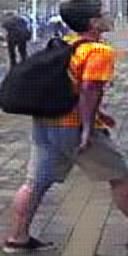} &
        \includegraphics[scale=0.14]{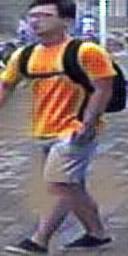} &
        \includegraphics[scale=0.14]{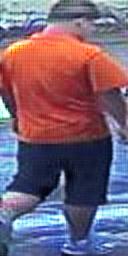} &
        \includegraphics[scale=0.14]{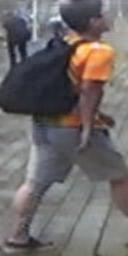} &
        \includegraphics[scale=0.14]{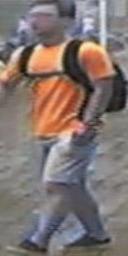} &
        \includegraphics[scale=0.14]{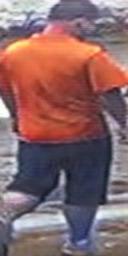} &
        \includegraphics[scale=0.14]{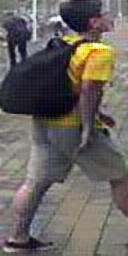} &
        \includegraphics[scale=0.14]{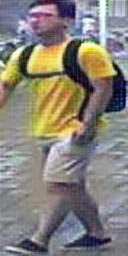} &
        \includegraphics[scale=0.14]{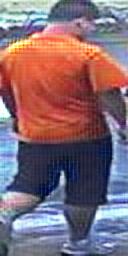} &

        \includegraphics[scale=0.14]{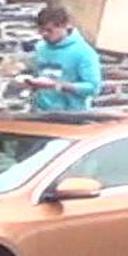} &
        \includegraphics[scale=0.14]{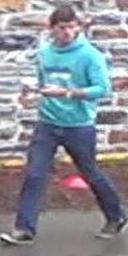} &
        \includegraphics[scale=0.14]{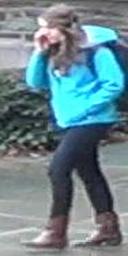} &
        \includegraphics[scale=0.14]{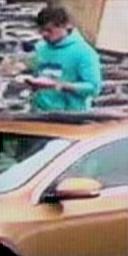} &
        \includegraphics[scale=0.14]{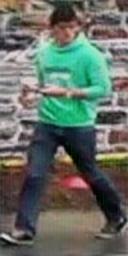} &
        \includegraphics[scale=0.14]{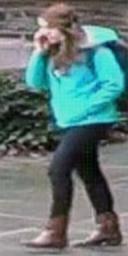} &
        \includegraphics[scale=0.14]{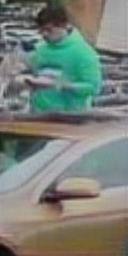} &
        \includegraphics[scale=0.14]{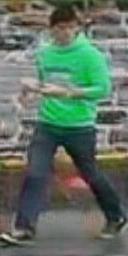} &
        \includegraphics[scale=0.14]{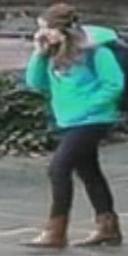} &
        \includegraphics[scale=0.14]{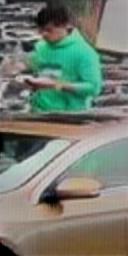} &
        \includegraphics[scale=0.14]{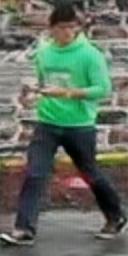} &
        \includegraphics[scale=0.14]{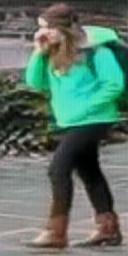} \\

         \multicolumn{3}{c}{\scriptsize Source images} & \multicolumn{3}{c}{\scriptsize CycleGAN} & \multicolumn{3}{c}{\scriptsize SPGAN} & \multicolumn{3}{c}{\scriptsize Ours} & \multicolumn{3}{c}{\scriptsize Source images} & \multicolumn{3}{c}{\scriptsize CycleGAN} & \multicolumn{3}{c}{\scriptsize SPGAN} & \multicolumn{3}{c}{\scriptsize Ours} \\
 \end{tabular}

    \captionof{figure}{ \label{fig:fig1}
(a)   {Although images of the same class are well gathered together after translation by existing domain translation-based methods \cite{deng2018image,wei2018person,deng2018similarity,chen2019instance},  the original intra-/inter-class distributions are disturbed.} The inter-sample relations of source-to-target translated images are not maintained. (b) Our proposed online relation regularization better preserves the inter-sample relations for  {generating informative training samples,  }improving cross-domain person re-ID. (c) Example triplets translated by CycleGAN \cite{zhu2017unpaired}, SPGAN \cite{deng2018image} and our proposed method.
Best viewed in color. }

\end{center}%

\end{figure*}

\begin{abstract}
Unsupervised domain adaptation (UDA) aims at adapting the model trained on a labeled source-domain dataset to an unlabeled target-domain dataset. The task of UDA on open-set person re-identification (re-ID) is even more challenging as the identities (classes) do not have overlap between the two domains. One major research direction was based on domain translation \cite{deng2018image,wei2018person,deng2018similarity,chen2019instance}, which, however, has fallen out of favor in recent years due to inferior performance compared to pseudo-label-based methods \cite{fan2018unsupervised,zhang2019self,yang2019selfsimilarity,ge2020mutual}. We argue that the domain translation has great potential on exploiting the valuable source-domain data but existing methods did not provide proper regularization on the translation process. Specifically, previous methods only focus on maintaining the identities of the translated images while ignoring the inter-sample relations during translation. To tackle the challenges, we propose an end-to-end structured domain adaptation framework with an online relation-consistency regularization term. During training, the person feature encoder is optimized to model inter-sample relations on-the-fly for supervising relation-consistency domain translation, which in turn, improves the encoder with informative translated images. The encoder can be further improved with pseudo labels, where the source-to-target translated images with ground-truth identities and target-domain images with pseudo identities are jointly used for training. In the experiments, our proposed framework is shown to achieve state-of-the-art performance on multiple UDA tasks of person re-ID. With the synthetic$\to$real translated images from our structured domain-translation network, we achieved second place in the Visual Domain Adaptation Challenge (VisDA) in 2020.
\end{abstract}


%

\section{Introduction}

%
%
%
%
\IEEEPARstart{P}{erson} re-identification (re-ID) \cite{9162561,8930088,8466034,wu2018exploit,lin2020unsupervised,wu2019progressive} aims at identifying images of the same person across multiple cameras.
Despite great advances of deep learning-based re-ID methods in recent years,
large domain gaps still pose great challenges on generalizing the trained models from a labeled source domain to an unlabeled target domain.
There are two main research directions towards solving the problem of domain adaptive person re-ID, \ie domain translation-based methods \cite{deng2018image,wei2018person,deng2018similarity,chen2019instance,tang2020cgan,huang2019sbsgan} and pseudo-label-based methods \cite{fan2018unsupervised,zhang2019self,yang2019selfsimilarity,ge2020mutual},
where the latter ones dominate the current literature with state-of-the-art performance.

Although domain translation-based methods have fallen out of favor in recent years due to their uncompetitive performance,
we argue that they have great potential to make full use of valuable source-domain data with accurate identity labels if they are imposed with proper regularization strategy.
Translating source-domain images to the target domain to create new training samples with identity labels is at the core of domain translation-based methods.
Previous works used identity-based regularization (\eg classification loss \cite{deng2018similarity,chen2019instance,zou2020joint}, contrastive loss \cite{deng2018image} or triplet loss \cite{tang2020cgan}) to preserve the ID-related appearance during image translation,
{\ie forcing different identities' images to be well separated from each other after translation.}
However, we observe that their domain-translated images cannot well maintain \emph{inter-sample relations} even with such ID-based constraints.
{The inter-sample relations in our paper are represented by semantic affinities/similarities between samples' encoded features in the latent space, rather than the simple positive/negative relations given by the ground-truth identity labels.}
 {As demonstrated in Figure \ref{fig:fig1}(a), in existing image-to-image translation frameworks, although images of the same class can still be well classified into the same class after translation, the inter-sample affinities in the latent space might not be well maintained to effectively regularize the translation process.}
 {For example, in the first case of Figure \ref{fig:fig1}(c), persons from two distinct identities are in yellow and orange, respectively. After domain translation by CycleGAN \cite{zhu2017unpaired} and SPGAN \cite{deng2018image}, the images of the different persons show similar color and appearances. The inter-sample relations are not well maintained.
In the second case, a male and a female are in the same color - blue. After translation by CycleGAN \cite{zhu2017unpaired}, the two images of the male change to green and blue, respectively.
After translation by SPGAN \cite{deng2018image},  the two images for the male are both in green while the girl is still in blue.
The inter-sample relations are not well maintained by the existing translation-based methods.}
 {We argue that maintaining inter-sample relations during image-to-image translation is critical for generating informative training samples, which cannot be effectively regularized by identity-based regularization in \cite{deng2018similarity,chen2019instance,zou2020joint,deng2018image,tang2020cgan}.}


To tackle this challenge, we propose an end-to-end structured domain adaptation (SDA) framework with a novel \emph{online relation-consistency regularization} term.
It consists of a structured domain-translation network, a source-domain person image encoder and a target-domain encoder.
The structured domain-translation network generates the source-to-target translated images, which can benefit the training of the target-domain encoder.
At the same time, two domains' encoders are coupledly trained to model inter-sample relations on-the-fly, which regularize the training of the domain-translation network.
We alternatively optimize the target-domain encoder and structured domain-translation network in each iteration, in order to exploit their mutual benefits.

More specifically, the structured domain-translation network adopts the CycleGAN \cite{zhu2017unpaired} architecture for translating source- and target-domain images.
A novel relation-consistency loss is proposed to regularize the training of source-to-target domain translation for maintaining the original inter-sample relations, which are generated by the source-domain encoder \textit{on-the-fly} (Figure \ref{fig:fig1}(b) and (c)).
Note that our relation-consistency loss is different from conventional ID-based constraints in \cite{deng2018image,wei2018person,deng2018similarity,chen2019instance,tang2020cgan}, which only apply  {``one-hot''} regularization that requires translated images of different classes to be well separated after translation  {but ignores the inter-sample relations}.

Meanwhile, the source-domain encoder is trained with source-domain images and ground-truth identities.
An improved target-domain encoder is trained with both the source-to-target translated images and target-domain images via a joint cross-domain label system, which is constructed with their associated ground-truth and pseudo labels. In this way, both the target-domain encoder and its generated pseudo labels can be improved with the optimization of the structured domain-translation network.
 {As shown in the first case of Figure \ref{fig:fig1}(c), the triplet translated by our method show distinguishable appearance for images of different persons. In the second case, the clothes of the male and the female are both translated to the same color (green). Our proposed method is able to effectively maintain their original inter-sample relations during translation via the proposed structured domain-translation network and relation-consistency loss.}

In summary, the contributions of this paper are three-fold:
\begin{itemize}
\item To properly exploit the valuable source-domain data in domain translation-based UDA methods,
we propose a novel online relation-consistency regularization term to better supervise the domain translation process.
\item The domain-translated images can serve as informative training samples to improve the target-domain encoder and help generate more accurate pseudo labels.
The domain-translation network and target-domain encoder alternately promote each other to achieve optimal re-ID performance.
\item Our framework achieves state-of-the-art performance on multiple domain adaptive person re-ID benchmarks.
Moreover, the proposed structured domain-translation method contributes to our solution \cite{ge2020improved} in the Visual Domain Adaptation Challenge (VisDA) in 2020, which ranks second out of 153 teams.
\end{itemize}

\section{Related Work}

\subsection{Methods for Unsupervised Domain Adaptation (UDA) on Person Re-ID}

There are two main categories for existing UDA methods on person re-ID, domain translation-based methods \cite{deng2018image,wei2018person,deng2018similarity,chen2019instance,8931652,tang2020cgan,huang2019sbsgan} and pseudo-label-based methods \cite{fan2018unsupervised,zhang2019self,yang2019selfsimilarity,ge2020mutual,8713423}.
This paper focuses on improving the former one, since we argue that the domain translation is effective to properly exploit the valuable source-domain images and their ground-truth identities, but existing works did not provide enough regularizations on the translation process.
Note that two kinds of methods are complementary to each other and can promote each other to achieve optimal performance.

\subsubsection{\textbf{Domain translation-based UDA methods for person re-ID}}

Domain translation-based methods \cite{deng2018image,wei2018person,deng2018similarity,chen2019instance,tang2020cgan,huang2019sbsgan} aimed at fine-tuning the target-domain re-ID model with source-to-target translated images and their ground-truth identities.
In order to preserve the original identities of the translated images,
ID-based regularizations were adopted on either pixel level \cite{wei2018person,huang2019sbsgan} or feature level \cite{deng2018image,deng2018similarity,chen2019instance,tang2020cgan} via the contrastive loss \cite{deng2018image}, classification loss \cite{deng2018similarity,chen2019instance,zou2020joint} or triplet loss \cite{tang2020cgan}.
Specifically, PTGAN \cite{wei2018person}  {and SBSGAN \cite{huang2019sbsgan}} imposed pixel-level constraints on maintaining the color consistency during the domain translation.
SPGAN \cite{deng2018image}  {and CGAN-TM \cite{tang2020cgan}} further maximized the feature-level similarities between translated images and the original ones.

Although the existing ID-based regularizations can somewhat preserve the original person appearance and the inter-identity images can be separated after translation,
they are too weak to properly maintain the original inter-sample relations and distributions of source-domain data during translation.
We argue that maintaining the inter-sample relations and the original distributions are critical for preserving the knowledge from the source domain and generating informative training samples, but unfortunately, ignored by existing works \cite{deng2018image,wei2018person,deng2018similarity,chen2019instance,huang2019sbsgan,tang2020cgan}.

\subsubsection{\textbf{Pseudo-label-based UDA methods for person re-ID}}

Pseudo-label-based methods \cite{fan2018unsupervised,song2018unsupervised,zhang2019self,yang2019selfsimilarity,ge2020mutual,yu2019unsupervised,zhong2019invariance}
achieved state-of-the-art performances by modeling relations between unlabeled target-domain data with generated pseudo labels, where a clustering-based pipeline was found effective.
PUL \cite{fan2018unsupervised}  {and UDAP \cite{song2018unsupervised}} first proposed a self-training scheme with clustering labels.
SSG \cite{yang2019selfsimilarity} and PAST \cite{zhang2019self} further extended this type of methods by introducing human part features and progressive training strategy.
Recently, MMT \cite{ge2020mutual} was proposed to adopt coupledly trained networks as well as their mean-teacher networks for mutual training, achieving state-of-the-art performances.

These methods generally focused on using only the target-domain data, and we found that they could be further improved by properly exploiting the valuable source-domain data with ground-truth identities, where domain translation has great potential.
Specifically, existing pseudo-label-based methods can be strengthened by jointly training with target-domain images and source-to-target translated images.

Note that the domain translation is crucial in jointly training, since the significant domain gaps may even harm the feature learning when directly training with two domains' raw images (see Table \ref{tab:ablation} ``Baseline'' v.s. ``Baseline + raw source-domain data'').
Our proposed structured domain-translation method can be well compatible with pseudo-label-based methods, including a clustering-based baseline \cite{song2018unsupervised} or state-of-the-art MMT \cite{ge2020mutual} (see Section \ref{sec:sota}).

\subsection{Generic Methods for Unsupervised Domain Adaptation}

Feature-level and pixel-level adaptations were commonly adopted by UDA methods for tackling more general tasks.
The feature-level adaptation methods \cite{long2015learning,sun2016deep,tzeng2017adversarial,das2018unsupervised,9057713,9115868} aimed at aligning the feature distributions between the source and target domains by learning domain-invariant features with a domain adversarial discriminator \cite{bousmalis2016domain,tzeng2017adversarial} or reducing the Maximum Mean Discrepancy (MMD) \cite{gretton2007kernel} distance between domains.
However,
such methods are unable to handle the open-set re-ID problem with disjoint label systems in two domains \cite{zhong2019invariance,panareda2017open,saito2018open}.

The other category of pixel-level adaptation methods \cite{hoffman2017cycada,li2019bidirectional,chen2019crdoco} minimized the domain shifts by translating images to the same domain,
which has been widely studied in semantic segmentation.
These pixel-level adaptation methods focused on the cross-domain class/prediction consistency, which are related to our method. However,
they still ignored the consistency of inter-sample relations and data distributions during translation, facing the same challenge as translated-based UDA methods for person re-ID \cite{deng2018image,wei2018person,deng2018similarity,chen2019instance}.

\subsection{Relation Preserving Embedding}

 {Relation preserving embedding \cite{roweis2000nonlinear,roweis2000nonlinear,shaw2009structure,tung2019similarity,park2019relational} aims at encoding features that preserve certain prior structure or relational information among samples.  There exist a large number of embedding techniques, which mainly include conventional matrix factorization-based methods \cite{roweis2000nonlinear,roweis2000nonlinear,shaw2009structure} and deep learning-based methods \cite{tung2019similarity,park2019relational}. Matrix factorization-based methods try to represent the inter-sample relations/affinities in the form of a matrix, such as adjacency matrix, Laplacian matrix, node transition probability matrix, Katz similarity matrix, \textit{etc}. However, those algorithms cannot be used as regularizations in our framework, as their objective functions are optimized via matrix factorization and do not support gradient-based optimization. Deep learning-based methods were also studied to capture non-linear structures among data points. However, these algorithms are different from our method in two aspects. First, those methods have different purposes from our method, \textit{i.e.} they target at directly encoding low-dimensional features to maintain certain inter-sample relations while we maintain inter-sample relations to regularize the translation network training. More importantly, they adopt different types of manually-specified inter-sample relations as learning targets while our inter-sample relational supervisions are online generated and are also alternatively optimized with the domain-translation network.}

\section{Structured Domain Adaptation}

\begin{figure}[t]
\centering
\includegraphics[width=1.0\linewidth]{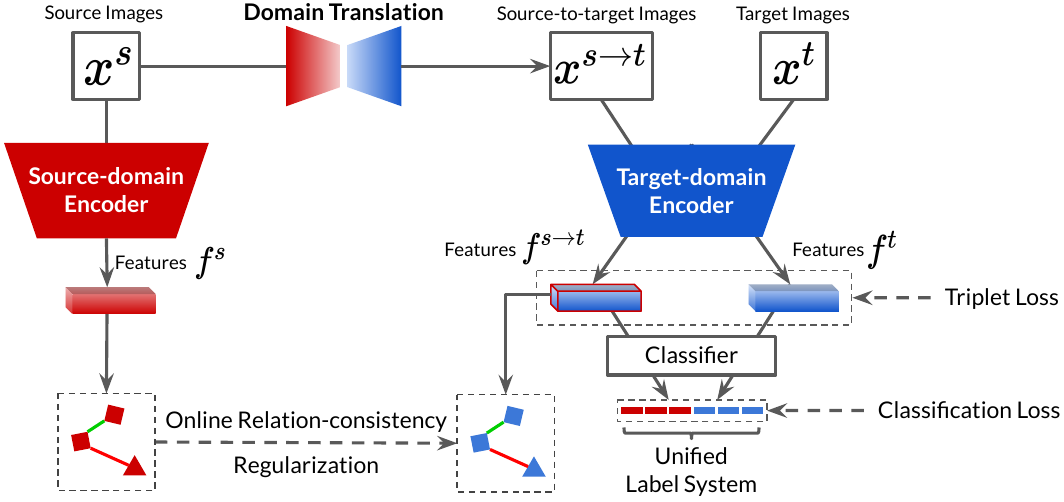}
\caption{Illustration of our structured domain adaptation (SDA) framework and the novel online relation-consistency regularization term.
The domain-translation network and the target-domain encoder alternately promote each other via joint training to achieve optimal re-ID performance. }
\label{fig:fm}
\end{figure}

We propose a structured domain adaptation framework with a novel online relation-consistency regularization term to tackle unsupervised domain adaptation (UDA) for person re-ID. The overall framework, as illustrated in Figure \ref{fig:fm}, consists of a structured domain-translation network and two domain-specific person image encoders. The translation network and target-domain encoder are jointly optimized and promote each other to learn more discriminative person features.

The key innovation of our framework lies in the generation of informative training samples by translating source-domain images into the target domain under the relation-consistency regularization generated by the image encoders on-the-fly.

\subsection{Source-domain Encoder Pre-training}
\label{sec:reid}

We pre-train the source-domain person image encoder $\mathcal{F}^s$
for (1) providing ``ground-truth'' inter-sample relations between source-domain images to regularize the proposed structured domain translation, and
(2) providing weight initialization for the target-domain person image encoder $\mathcal{F}^t$.

Given source-domain samples $\sX^s$, the encoder $\mathcal{F}^s$ is trained to transform each sample $x^s \in \sX^s$ into a feature vector $\vf^s=\mathcal{F}^s(x^s)$.
If the feature vector $\vf^s$ is properly embedded, it could be used to correctly predict its ground-truth identity $y^s$ with a learnable classifier $\mathcal{C}^s: \vf^s \to \{1,\cdots,p^s \}$, where $p^s$ is the number of identities in the source domain.
A cross-entropy classification loss $\ell_{ce}$ and a triplet loss \cite{hermans2017defense} are adopted jointly for training,
{
\begin{align}
\label{eq:source}
\mathcal{L}_\text{enc}^s(\mathcal{F}^s,&\mathcal{C}^s) =
\mathbb{E}_{x^s \sim \sX^s}\left[ \ell_\text{ce} ( \mathcal{C}^s(\vf^s), y^s  )\right] \nonumber \\
&+
\mathbb{E}_{x^s \sim \sX^s}\left[
(\|\vf^s-\vf^s_p\| + m
 - \|\vf^s-\vf^s_n\| )^+\right],
\end{align}}%
where
$(\cdot)^+=\max(0,\cdot)$ with a margin $m$, and the subscripts $_p, _n$ denote the mini-batch's hardest positive and negative feature indexes for the anchor $\vf^s$.

Once trained, $\mathcal{F}^s$ is frozen to provide stable regularizations for inter-sample relations.

\subsection{Structured Domain Translation}
\label{sec:sdt}

We propose a structured domain-translation (SDT) network to generate informative training samples by translating source-domain images $\sX^s$ to the target domain, which focuses not only on image-style transfer but more on how to maintain their original inter-sample relations and distributions.
We adopt the widely-used CycleGAN~\cite{zhu2017unpaired} architecture for our translation network, which is trained to translate images along two directions
with corresponding generators $\mathcal{G}^{s\to t}$ and $\mathcal{G}^{t\to s}$.

\subsubsection{\textbf{Conventional cycle generation losses}}
The general training objective of a CycleGAN \cite{zhu2017unpaired} for image-to-image translation consists of the adversarial losses $\mathcal{L}_\text{adv}^s$, $\mathcal{L}_\text{adv}^t$, the cyclic reconstruction loss $\mathcal{L}_\text{cyc}$ and the appearance consistency loss $\mathcal{L}_\text{apr}$. We adopt the loss function of LSGAN \cite{mao2017least} with two domain discriminators $\mathcal{D}^s$ and $\mathcal{D}^t$ as
{
\begin{align}
\mathcal{L}_\text{adv}^s(\mathcal{G}^{t\to s},\mathcal{D}^s)&=\mathbb{E}_{x^s\sim \sX^s}\left[\mathcal{D}^s(x^s)^2 \right] \nonumber \\
&+\mathbb{E}_{x^t\sim \sX^t}\left[\left(\mathcal{D}^s(\mathcal{G}^{t\to s}(x^t))-1 \right)^2 \right],
\nonumber \\
\mathcal{L}_\text{adv}^t(\mathcal{G}^{s\to t},\mathcal{D}^t)&=\mathbb{E}_{x^t\sim \sX^t}\left[\mathcal{D}^t(x^t)^2 \right] \nonumber \\
&+\mathbb{E}_{x^s\sim \sX^s}\left[\left(\mathcal{D}^t(\mathcal{G}^{s\to t}(x^s))-1 \right)^2 \right].
\end{align}}%
The cyclic reconstruction loss supervises the pixel-level generation by translating the images along two directions twice,
{
\begin{align}
\mathcal{L}_\text{cyc}(\mathcal{G}^{s\to t}, &\mathcal{G}^{t\to s}) = \mathbb{E}_{x^s\sim \sX^s}\left[\| \mathcal{G}^{t\to s}(\mathcal{G}^{s\to t}(x^s)) - x^s \|_1 \right] \nonumber \\
&+\mathbb{E}_{x^t\sim \sX^t}\left[\| \mathcal{G}^{s\to t}(\mathcal{G}^{t\to s}(x^t)) - x^t \|_1 \right] .
\end{align}}%
The appearance consistency loss \cite{taigman2017unsupervised} help maintain the general color composition after domain translation,
{
\begin{align}
\mathcal{L}_\text{apr}(\mathcal{G}^{s\to t}, \mathcal{G}^{t\to s})&=\mathbb{E}_{x^s\sim \sX^s}\left[\| \mathcal{G}^{t\to s}(x^s) - x^s \|_1 \right] \nonumber \\
&+
\mathbb{E}_{x^t\sim \sX^t}\left[\| \mathcal{G}^{s\to t}(x^t) - x^t \|_1 \right].
\end{align}}%

Despite the fact that the above loss terms guide the source-domain images to have target-domain image style and the appearance consistency loss  $\mathcal{L}_\text{apr}$ can preserve the person appearance to some extent,
the generated images generally fail to maintain their original identities, where inter-class images cannot be well separated after translation.
To tackle the problem, existing domain translation-based methods adopts ID-based constraints to regularize the translation process, \eg contrastive loss \cite{deng2018image}, classification loss \cite{deng2018similarity,chen2019instance,zou2020joint} or triplet loss \cite{tang2020cgan}.

However, we observe that ``one-hot'' ID-based regularizations adopted by existing works \cite{deng2018image,wei2018person,deng2018similarity,chen2019instance,zou2020joint,tang2020cgan}  {only require the translated images to be well separated according to their identities, but} are not able to preserve the accurate inter-sample structures as well as the original source-domain data distributions (Figure \ref{fig:fig1}(c)).
As a result, the translated images are not accurate enough to optimize the target-domain image encoder.
We argue that properly maintaining the original inter-sample relations is crucial to the domain translation.

\subsubsection{\textbf{Online relation-consistency regularization}}
Since there are no ``ground-truth'' inter-sample relations,
we propose to use the pre-trained $\mathcal{F}^s$ to provide relation supervision on-the-fly for effectively regularizing the translation.
Intuitively, inter-sample relations can be modeled by the ratio of feature similarities between images.
In person re-ID, we found that triplets with intra-/inter-identity samples are generally representative and can be used for modeling inter-sample relations.

Given a source-domain image $x^s$, its positive sample $x^s_p$ with the same identity, and its negative sample $x^s_n$ with a different identity,
we can measure their similarity-based inter-sample relations on-the-fly among the triplet with a softmax-like function,
{
\begin{align}
\mathcal{R}(x^s; \mathcal{F}^s)=\frac{\exp \langle \vf^s, \vf^s_p\rangle}
{\exp \langle \vf^s, \vf^s_p\rangle + \exp \langle \vf^s, \vf^s_n \rangle} \in [0,1),
\label{eq:rel_s}
\end{align}}%
where $\vf^s$, $\vf_p^s$, $\vf_n^s$ are the features encoded by the pre-trained source-domain encoder $\mathcal{F}^s$ on the image samples $x^s$, $x_p^s$, $x_n^s$, respectively, and $\langle \cdot, \cdot \rangle$ is the inner product between two feature vectors to measure their similarity.
Similar to \cite{hermans2017defense}, we utilize only the most difficult triplet of each sample $x^s$ within a batch, \ie the hardest positive ($\vf^s_p$) and negative ($\vf^s_n$) samples for each $\vf^s$. Note that $\mathcal{R}(x^s; \mathcal{F}^s)$ is a \emph{continuous} value in $[0,1)$ to measure the ratio of pairwise similarities.

After translating source-domain images to the target domain by $\mathcal{G}^{s \to t}$, we obtain the features of the source-to-target translated triplet
(${\vf}^{s \to t}, {\vf}^{s \to t}_p, {\vf}^{s \to t}_n$), which are encoded by the target-domain encoder $\mathcal{F}^t$ (to be discussed in Section \ref{sec:joint}).
Similarly, the continuous similarity ratio in $[0,1)$ between the translated images can also be measured by a softmax-like function as
{
\begin{align}
\mathcal{R}&(x^s; \mathcal{G}^{s \to t}, \mathcal{F}^t) \nonumber \\
&= \frac{\exp \langle {\vf}^{s \to t}, {\vf}^{s \to t}_p \rangle}
{\exp \langle {\vf}^{s \to t}, {\vf}^{s \to t}_p\rangle  + \exp \langle {\vf}^{s \to t}, {\vf}^{s \to t}_n\rangle} \in [0,1).
\label{eq:rel_T}
\end{align}}%

We claim that, if the domain-translation network well preserves the source-domain images' inter-sample relations, their softmax-triplet responses before and after the translation should be as close as possible.
Based on this assumption, a novel relation-consistency loss is introduced to regularize the inter-sample relations after the source-to-target translation. The loss is modeled as a ``soft'' binary cross-entropy loss,
{
\begin{align}
\label{eq:sc}
\mathcal{L}_\text{rc}&(\mathcal{G}^{s\to t}) \nonumber \\
&=
\mathbb{E}_{x^s\sim \sX^s} \left[ \ell_\text{bce} \left( \mathcal{R}(x^s; \mathcal{G}^{s \to t}, \mathcal{F}^t),  \mathcal{R}(x^s; \mathcal{F}^s) \right) \right],
\end{align}}%
where $\ell_\text{bce}(p, q) = - q \log p - \left(1- q\right) \log \left( 1-p\right)$ with the soft label $q \in [1,0)$. In other words, we use the source-domain inter-sample relations $\mathcal{R}(x^s; \mathcal{F}^s)$ as \emph{soft} learning targets for supervising the translated inter-sample relations $\mathcal{R}(x^s; \mathcal{G}^{s \to t}, \mathcal{F}^t)$.

We have also investigated alternative designs of our relation-consistency loss for measuring the inter-sample relations and regularizing the translation process, which will be further discussed in Section \ref{sec:alter}.

\subsubsection{\textbf{Differences from existing ID-based regularizations}}
There are two key differences between the proposed online relation-consistency loss (Eq. (\ref{eq:sc})) and ID-based regularizations in existing works, \eg classification loss \cite{deng2018similarity,chen2019instance,zou2020joint}, contrastive loss \cite{deng2018image} or triplet loss.
\begin{itemize}
\item Existing ID-based regularizations only require the samples from different classes to be well separated after translation, which is too weak to maintain inter-sample relations and their original data distributions. As long as the translated samples are classified correctly, even if they did not maintain inter-sample relations well, they receive little penalty. In contrast, our proposed online regularization aims at maintaining continuous and more sensitive relation measurements (Eq. (\ref{eq:rel_s})) during domain translation.
\item Existing regularizations utilize static learning targets (identity labels), while our regularization term generates relation measurements with the current image encoders on-the-fly to provide adaptive supervisions.
In other words, previous ones only acquire knowledge from ground-truth labels, while ours exploits valuable knowledge from both ground-truth labels and the pre-trained source-domain encoder.
\end{itemize}

Besides the domain translation-based UDA methods as mentioned above, there exist some works \cite{liu2018pose,ge2018fd,zheng2019joint} which leveraged generative models on fully-supervised person re-ID tasks. They also focused on preserving person identities with ID-based regularizations, which are too weak to maintain inter-sample relations as discussed above.

\subsubsection{\textbf{Joint training objective}}

During training, we fix $\mathcal{F}^s$ and alternately update $\mathcal{F}^t$ and SDT in each iteration to avoid bias amplification,
where the SDT network is optimized with
{
\begin{align}
\mathcal{L}_\text{sdt}&(\mathcal{G}^{s\to t}, \mathcal{G}^{t\to s}, \mathcal{D}^s, \mathcal{D}^t) =
\lambda_\text{rc}\mathcal{L}_\text{rc}(\mathcal{G}^{s\to t}) \nonumber \\
&+\lambda_\text{cyc}\mathcal{L}_\text{cyc}(\mathcal{G}^{s\to t}, \mathcal{G}^{t\to s})
+\lambda_\text{apr}\mathcal{L}_\text{apr}(\mathcal{G}^{s\to t}, \mathcal{G}^{t\to s}) \nonumber \\
&+\lambda_\text{adv}\left(\mathcal{L}_\text{adv}^s(\mathcal{G}^{t\to s}, \mathcal{D}^s)+\mathcal{L}_\text{adv}^t(\mathcal{G}^{s\to t}, \mathcal{D}^t)\right).
\label{eq:sdt}
\end{align}}%
Here $\lambda_\text{rc}$, $\lambda_\text{cyc}$, $\lambda_\text{adv}$ and $\lambda_\text{apr}$ are the weighting factors for different loss terms.

\subsection{Target-domain Encoder with Translated Images}
\label{sec:joint}

\begin{algorithm}[tb]
\caption{Structured domain adaptation for person re-ID}
\label{overallalg}
\footnotesize
\begin{algorithmic}[1]
\REQUIRE Labeled source-domain data $\sX^s$, unlabeled target-domain data $\sX^t$;
\REQUIRE Weighting factors $\lambda_\text{rc},\lambda_\text{cyc},\lambda_\text{adv},\lambda_\text{apr}$ for Eq. (\ref{eq:sdt});
\STATE Pre-train source-domain encoder $\mathcal{F}^s$ by minimizing Eq. (\ref{eq:source}) on $\sX^s$;
\STATE  {Initialize target-domain encoder $\mathcal{F}^t$ by loading the weights of $\mathcal{F}^s$};

\FOR{$n$ in $[1, \text{num\_epochs}]$}
\STATE Create pseudo labels by clustering $\mathcal{F}^t(\sX^t)$;
\FOR{each mini-batch $\sB^s \subset \sX^s$, $\sB^t \subset \sX^t$}
\STATE Translate $\sB^s$ into the target domain as $\sB^{s\to t}$ by $\mathcal{G}^{s\to t}$; \
\STATE Update $\mathcal{G}^{s\to t}, \mathcal{G}^{t\to s}$ by minimizing the objective function Eq. (\ref{eq:sdt}) with $\mathcal{D}^s, \mathcal{D}^t$ fixed, where the inter-sample relations are measured by $\mathcal{F}^s$ and $\mathcal{F}^t$ on-the-fly; \
\STATE Update $\mathcal{F}^t$ by minimizing the objective function Eq. (\ref{eq:pic}) with $\sB^{s\to t} \cup \sB^t$; \
\STATE Update $\mathcal{D}^s, \mathcal{D}^t$ by maximizing the objective function Eq. (\ref{eq:sdt}) with $\mathcal{G}^{s\to t}, \mathcal{G}^{t\to s}$ fixed. \
\ENDFOR
\ENDFOR
\end{algorithmic}
\end{algorithm}

For training the target-domain encoder $\mathcal{F}^t$ in our framework,
arbitrary pseudo-label-based methods (\eg UDAP \cite{song2018unsupervised}, MMT \cite{ge2020mutual}) can be adopted as a baseline, and we can improve them by jointly training two domains' images, \ie the source-to-target images translated by our SDT and raw target-domain images.
Specifically,
we can create a unified training image set $\sX=\sX^{s\to t} \cup \sX^t$ with a unified label set to supervise a cross-domain identity classifier $\mathcal{C}^t:\vf\rightarrow\{1,\cdots, p^s+\hat{p}^t \}$,
where both the labeled source-to-target translated images $\sX^{s\to t}$ and the pseudo-labeled target-domain images $\sX^{t}$ serve as informative training samples with non-overlapping real or pseudo identity labels.

Note that the pseudo label creation is a general pipeline in UDA tasks and is not the focus of our method.
Our framework can also work without pseudo labels (to be discussed in Section \ref{sec:com_dt_1}), \ie only the labeled source-to-target translated images are adopted for training the target-domain encoder.

Here, we take  {the modified version of} a clustering-based baseline \cite{song2018unsupervised} as an example.  {Note that the original version of \cite{song2018unsupervised} only adopted the triplet loss for training, while our modified version adopts both the cross-entropy classification loss and triplet loss to achieve better performance.}
Target-domain data $\sX^t$'s encoded features $\{\vf^t\}$ are clustered into $\hat{p}^t$ classes and images within the same cluster are assigned the same pseudo label.
The target-domain encoder ${\mathcal{F}^t}$ can then be trained in a fully-supervised manner.
Specifically,
each sample $x \in \sX$ is assigned a corresponding label $y \in \{1,\cdots,p^s+\hat{p}^t\}$, and
${\mathcal{F}^t}$ is optimized with the objective function similar to source-domain encoder learning in Eq. (\ref{eq:source}) but with a joint-domain label set,
{
\begin{align}
\label{eq:pic}
\mathcal{L}_\text{enc}^t(\mathcal{F}^t,\mathcal{C}^t) &=
\mathbb{E}_{x \sim \sX}\left[ \ell_\text{ce} ( \mathcal{C}^t(\vf), y  )\right] \nonumber\\
&+
\mathbb{E}_{x \sim \sX}\left[
(\|\vf-\vf_p\| + m
 - \|\vf-\vf_n\| )^+\right].
\end{align}}%

The target-domain encoder $\mathcal{F}^t$ can therefore take full advantages of
(1) the source-to-target images translated by our $\mathcal{G}^{s\to t}$, which better maintains their inter-sample relations, and
(2) the unified cross-domain label set that consists of both the valuable ground-truth source-domain identity labels and the target-domain pseudo labels.
$\mathcal{F}^t$ trained by this strategy is shown to encode more discriminative features for distinguishing target-domain identities.

In our overall framework, the source-domain encoder $\mathcal{F}^s$ is fixed after pre-training, and the structured domain-translation (SDT) network and the target-domain encoder $\mathcal{F}^t$ alternately promote each other via joint training to achieve optimal re-ID performance.
When fixing $\mathcal{F}^t$,
it measures translated inter-sample relations (Eq. (\ref{eq:rel_T})) for regularizing SDT via
$\mathcal{L}_\text{rc}$ (Eq. (\ref{eq:sc})).
When fixing SDT, it generates training samples $\sX^{s\to t}$ to optimize $\mathcal{F}^t$ (Eq. (\ref{eq:pic})).
Once $\mathcal{F}^t$ is further trained to achieve better re-ID performance on the target domain, it could in turn generate more accurate pseudo labels and measure more accurate inter-sample relations for further improving SDT.

After training, only $\mathcal{F}^t$ is used to encode target-domain samples into features for pairwise similarity estimation without extra parameters and computational costs.
The overall algorithm is summarized in Algorithm \ref{overallalg}.

\section{Experiments}

\subsection{Datasets and Evaluation Metric}

We evaluate our framework on four real$\to$real adaptation tasks for person re-ID\footnote{Code for general datasets is available at \url{https://github.com/SDA-ReID/SDA}.}, including DukeMTMC-reID$\to$Market-1501, Market-1501$\to$DukeMTMC-reID, Market-1501$\to$MSMT17 and DukeMTMC-reID$\to$MSMT17 following the experimental setup in state-of-the-arts \cite{ge2020mutual,yang2019selfsimilarity,zhang2019self}.
\begin{itemize}
\item DukeMTMC-reID \cite{dukemtmc}
contains 36,411 images of 702 identities for training and another 702 identities for testing, with all the images captured from 8 cameras.
\item Market-1501 \cite{market} consists of 12,936 images of 751 identities for training and
19,281 images of 750 identities for testing, which are shot by 6 cameras.
\item MSMT17 \cite{wei2018person} is the most challenging dataset with 126,441 images of 4,101 identities from 15 cameras, where 1,041 identities are used for training.
\end{itemize}

Besides the widely-used real$\to$real benchmarks above, we also adopt the proposed structured domain-translation method in the synthetic$\to$real adaptation task of the Visual Domain Adaptation Challenge (VisDA-2020)\footnote{Code for VisDA 2020 challenge is available at \url{https://github.com/yxgeee/VisDA-ECCV20}.}.




\begin{itemize}
\item The synthetic source-domain dataset PersonX \cite{sun2019dissecting} is generated based on Unity \cite{riccitiello2015john} engine, containing 20,280 images out of 700 identities captured by 6 cameras. The target-domain dataset consists of real-world images captured from 5 cameras, \ie 13,198 images for training, 377 images for the query of target\_val, 3,600 images for the gallery of target\_val, 1,578 images for the query of target\_test and 24,006 images for the gallery of target\_test.  {Only the train set for the real-world images are used for training. The target\_val is used for evaluation offline and the target\_test is used for evaluation online. }
\end{itemize}

 {Note that both the train and test sets for source and target domains are required in our experiments, where only train sets are used for model optimization and the test sets are used for model evaluation. Specifically, for source-domain pre-training, the source-domain train set is used for training and the source-domain test set is used for evaluation. For our structured domain-translation training, the train sets of both source and targets domains are used for training, and the target-domain test set is used for evaluation.}
Mean Average Precision (mAP) and Cumulative Matching Characteristics (CMC) accuracies are utilized to test the methods performance.

\subsection{Implementation Details}
\label{sec:imp}

\subsubsection{\textbf{Network architecture}}
We adopt ResNet-50 \cite{he2016deep} as the backbone for the source-domain and target-domain person image encoders, which are initialized with ImageNet-pretrained \cite{deng2009imagenet} weights.
 {We adopt the CycleGAN \cite{zhu2017unpaired} architecture for our generators $\mathcal{G}^{s\to t}$, $\mathcal{G}^{t\to s}$ and the PatchGAN \cite{isola2017image} architecture for our discriminators $\mathcal{D}^s$, $\mathcal{D}^t$. Specifically, each generator consists of three convolution-IN-ReLU blocks, nine residual blocks \cite{he2016deep},  two deconvolution-IN-ReLU blocks and the last one convolution mapping feature maps to RGB images.}
The target-domain encoder $\mathcal{F}^t$ and structured domain-translation network are alternately updated in each iteration to avoid unstable training.
Furthermore, we adopt a momentum encoder \cite{he2019momentum} (denoted as $\mathcal{F}^t_*$)
to replace $\mathcal{F}^t$ in Eq. (\ref{eq:rel_T}) for measuring more stable triplet relations after domain translation.
In particular,
we denote the parameters of $\mathcal{F}^t$ and $\mathcal{F}^t_*$ as $\theta^{(T)}$ and $\theta_*^{(T)}$ at iteration $T$.
$\theta_*^{(T)}$ can be calculated as $\theta_*^{(T)}=\alpha\theta_*^{(T-1)}+(1-\alpha)\theta^{(T)}$, where $\theta_*^{(0)}=\theta^{(0)}$ and $\alpha=0.999$ is the momentum coefficient.
Intuitively, the momentum encoder could provide more reliable inter-sample relations for regularizing the Structured Domain Translation (SDT) since it eases the training bias caused by unstable translation results.

\subsubsection{\textbf{Training data organization}}
 {We adopt a $PK$ sampler for training, \textit{i.e.} mini-batch = $P$ identities $\times$ $K$ samples. For source-domain pre-training, each mini-batch contains 56 source-domain images of $P = 8$ ground-truth identities ($K = 7$ for each identity). When jointly training the structured domain-translation network and target-domain encoder,}
each mini-batch contains 56 source-domain images of 8 ground-truth identities and 56 target-domain images of 8 pseudo identities.
The pseudo identities are assigned by clustering algorithm and updated before each epoch.
 {With such a $PK$ sampler, the training samples for two domains can be well balanced even the two domains' datasets differ a lot regarding their scales.}
All images are resized to 256$\times$128.
Randomly erasing \cite{zhong2017random}, cropping and flipping are applied to each image.

\subsubsection{\textbf{Network optimization}}
ADAM optimizer is adopted to optimize the networks with weighting factors $\lambda_\text{rc}=1$, $\lambda_\text{adv}=1$, $\lambda_\text{cyc}=10$, $\lambda_\text{apr}=0.5$ and triplet margin $m=0.3$.
The initial learning rates ($lr$) are set to 0.00035 for person image encoders and 0.0002 for the SDT network.
The source-domain pre-training iterates for 30 epochs and the learning rate decreases to 1/10 of its previous value every 10 epochs.
The proposed joint training scheme (Algorithm \ref{overallalg}) iterates for 50 epochs,
where the learning rate is constant for the first 25 epochs and then gradually decreases to 0 for another 25 epochs following the formula $lr=lr\times(1.0-\max(0,epoch-25)/25)$.

\begin{table*}[t]
	\footnotesize
	\centering
	\caption{Unsupervised domain adaptation performances by state-of-the-art methods and our proposed SDA on person re-ID datasets, \eg DukeMTMC-reID~\cite{dukemtmc}, Market-1501~\cite{market}.}
	\vspace{-5pt}
	\label{tab:sota}
	\begin{center}
	\begin{tabular}{P{5cm}|C{1cm}C{1cm}C{1cm}C{1cm}|C{1cm}C{1cm}C{1cm}C{1cm}}
	\hline
	\multicolumn{1}{c|}{\multirow{2}{*}{Methods}} & \multicolumn{4}{c|}{DukeMTMC-reID$\to$Market-1501} & \multicolumn{4}{c}{Market-1501$\to$DukeMTMC-reID} \\
	\cline{2-9}
	\multicolumn{1}{c|}{} & mAP & top-1 & top-5 & top-10 & mAP & top-1 & top-5 & top-10 \\
	\hline \hline
	\multicolumn{9}{l}{{\textit{Domain translation-based UDA methods}}} \\
	\hline
    PTGAN~\cite{wei2018person} (CVPR'18) &  - & 38.6 &-&-& - & 27.4& -& - \\
    SPGAN~\cite{deng2018image} (CVPR'18) & 22.8 & 51.5 & 70.1 & 76.8 & 22.3 & 41.1 & 56.6 & 63.0 \\
        CR-GAN~\cite{chen2019instance} (ICCV'19) & 29.6 & 59.6 & - & - & 30.0 & 52.2 & - & - \\
         {SBSGAN~\cite{huang2019sbsgan} (ICCV'19)} & 27.3 & 58.5 & - & - & 30.8 & 53.5 & - & - \\
	\hline
	\multicolumn{9}{l}{{\textit{Pseudo-label-based UDA methods \& Others}}} \\
	\hline
    PUL~\cite{fan2018unsupervised} (TOMM'18) & 20.5 & 45.5 & 60.7 & 66.7 & 16.4 & 30.0 & 43.4 & 48.5 \\
    TJ-AIDL~\cite{wang2018transferable} (CVPR'18) & 26.5 & 58.2 & 74.8 & 81.1 & 23.0 & 44.3 & 59.6 & 65.0 \\
    HHL~\cite{zhong2018generalizing} (ECCV'18) & 31.4 & 62.2 & 78.8 & 84.0 & 27.2 & 46.9 & 61.0 & 66.7 \\
    CFSM~\cite{chang2018disjoint} (AAAI'19) & 28.3 & 61.2 & - & - & 27.3 & 49.8 &- & -  \\
    BUC~\cite{lin2019aBottom} (AAAI'19) & 38.3 & 66.2 & 79.6 & 84.5 & 27.5 & 47.4 & 62.6 & 68.4 \\
    ARN~\cite{li2018adaptation} (CVPR'18-WS) & 39.4 & 70.3 & 80.4 & 86.3 & 33.4 & 60.2 & 73.9 & 79.5 \\
    ECN~\cite{zhong2019invariance} (CVPR'19) & 43.0 & 75.1 & 87.6 & 91.6 & 40.4 & 63.3 & 75.8 & 80.4 \\
    UCDA~\cite{qi2019novel} (ICCV'19) & 30.9 & 60.4 & - & - & 31.0 & 47.7 & - & - \\
    PDA-Net~\cite{li2019cross} (ICCV'19) & 47.6 & 75.2 & 86.3 & 90.2 & 45.1 & 63.2 & 77.0 & 82.5 \\
    PCB-PAST~\cite{zhang2019self} (ICCV'19) & 54.6 & 78.4 & - & - & 54.3 & 72.4 & - & - \\
    SSG~\cite{yang2019selfsimilarity} (ICCV'19) & 58.3 & 80.0 & 90.0 & 92.4 & 53.4 & 73.0 & 80.6 & 83.2 \\
    ECN++~\cite{zhong2020learning} (TPAMI'20) & 63.8 & 84.1 & 92.8 & 95.4 & 54.4 & 74.0 & 83.7 & 87.4 \\
   MMCL~\cite{wang2020unsupervised} (CVPR'20) & 60.4 & 84.4 & 92.8 & 95.0 & 51.4 & 72.4 & 82.9 & 85.0 \\
   SNR~\cite{jin2020style} (CVPR'20) & 61.7 & 82.8 & - & - & 58.1 & 76.3 & - & - \\
  AD-Cluster~\cite{zhai2020ad} (CVPR'20) & 68.3 & 86.7 & 94.4 & 96.5 & 54.1 & 72.6 & 82.5 & 85.5 \\
  DG-Net++~\cite{zou2020joint} (ECCV'20) & 61.7 & 82.1 & 90.2 & 92.7 & 63.8 &  78.9 & 87.8 & 90.4 \\
  ADTC \cite{jiattention}  (ECCV'20) &  59.7 & 79.3 & 90.8 & 94.1 & 52.5 & 71.9 & 84.1 & 87.5 \\
  D-MMD \cite{mekhazni2020unsupervised} (ECCV'20) & 48.8 & 70.6 & 87.0 & 91.5 &46.0 & 63.5 & 78.8  & 83.9 \\
  JVTC \cite{li2020joint} (ECCV'20) &   61.1 & 83.8 & 93.0 & 95.2 &  56.2 & 75.0 & 85.1 & 88.2 \\
  GPR \cite{luogeneralizing} (ECCV'20) & {71.5}& {88.1} &  {94.4} & {96.2} &  {65.2} & {79.5} & {88.3} & {91.4} \\
  NRMT \cite{zhao2020unsupervised} (ECCV'20) &  {71.7} & {87.8} & {94.6} & {96.5} & 62.2 & 77.8 & 86.9 & 89.5 \\
  DCML \cite{chendeep} (ECCV'20) & {72.3} & {88.2} & {94.9} & {96.4} & 63.5 & 79.3 & 86.7 & 89.5 \\
  GDS-H \cite{jin2020global} (ECCV'20) & 61.2 & 81.1 & - & - & 55.1 & 73.1 & - & -\\
    \hline
    UDAP~\cite{song2018unsupervised} w/ {DBSCAN} (arXiv'18, PR'20) & 53.7 & 75.8 & 89.5 & 93.2 & 49.0 & 68.4 & 80.1 & 83.5 \\
     {UDAP~\cite{song2018unsupervised} w/ {DBSCAN} (modified)} & 59.0 & 80.7 & 90.5 & 93.4 & 50.1 & 68.2 & 79.2 & 82.7 \\
	Our SDA w/ {DBSCAN} & \textbf{70.0} & \textbf{86.9} & \textbf{94.4} & \textbf{96.3} & \textbf{61.4} & \textbf{76.5} & \textbf{86.6} & \textbf{89.7} \\
	Our SDA w/ {$k$-means} & {66.4} & {86.4} & {93.1} & {95.6} & {56.7} & {74.0} & {84.1} & {87.7} \\
	\hline
	MMT~\cite{ge2020mutual} w/ {$k$-means} (ICLR'20) & {71.2} & {87.7} & {94.9} & {96.9} & {65.1} & {78.0} & {88.8} & {92.5} \\
	 {SPGAN~\cite{deng2018image} + MMT~\cite{ge2020mutual} w/ {$k$-means}} & 68.5 & 86.9 & 93.2 & 96.0 & 63.7 & 76.0 & 86.2 & 90.5 \\
	Our SDA + MMT~\cite{ge2020mutual} w/ {$k$-means} & \textbf{74.3} & \textbf{89.7} & \textbf{95.9} & \textbf{97.4} & \textbf{66.7} & \textbf{79.9} & \textbf{89.1} & \textbf{92.7} \\
	\hline
	\end{tabular}
	\end{center}
\end{table*}

\begin{table*}[t]
	\footnotesize
	\centering
	\caption{Unsupervised domain adaptation performances by state-of-the-art methods and our proposed SDA on MSMT17~\cite{wei2018person}.}
	\vspace{-5pt}
	\label{tab:sota1}
	\begin{center}
	\begin{tabular}{P{5cm}|C{1cm}C{1cm}C{1cm}C{1cm}|C{1cm}C{1cm}C{1cm}C{1cm}}
	\hline
	\multicolumn{1}{c|}{\multirow{2}{*}{Methods}} & \multicolumn{4}{c|}{Market-1501$\to$MSMT17} & \multicolumn{4}{c}{DukeMTMC-reID$\to$MSMT17} \\
	\cline{2-9}
	\multicolumn{1}{c|}{} & mAP & top-1 & top-5 & top-10 & mAP & top-1 & top-5 & top-10 \\
	\hline \hline
	\multicolumn{9}{l}{{\textit{Domain translation-based UDA methods}}} \\
	\hline
    PTGAN~\cite{wei2018person} (CVPR'18) & 2.9 & 10.2 & - & 24.4 & 3.3 & 11.8 & - & 27.4 \\
    \hline
	\multicolumn{9}{l}{{\textit{Pseudo-label-based UDA methods \& Others}}} \\
	\hline
    ECN~\cite{zhong2019invariance} (CVPR'19) & 8.5 & 25.3 & 36.3 & 42.1 & 10.2 & 30.2 & 41.5 & 46.8 \\
    SSG~\cite{yang2019selfsimilarity} (ICCV'19) & 13.2 & 31.6 &- & 49.6 & 13.3 & 32.2 & - & 51.2 \\
    ECN++~\cite{zhong2020learning} (TPAMI'20) & 15.2 & 40.4 & 53.1 & 58.7 & 16.0 & 42.5 & 55.9 & 61.5 \\
   MMCL~\cite{wang2020unsupervised} (CVPR'20) & 15.1 & 40.8 & 51.8 & 56.7 & 16.2 & 43.6 & 54.3 & 58.9 \\
   DG-Net++~\cite{zou2020joint} (ECCV'20) & {22.1} & {48.4} & {60.9} & {66.1} & {22.1} & {48.8} & {60.9} & {65.9} \\
   D-MMD \cite{mekhazni2020unsupervised} (ECCV'20) & 13.5 & 29.1 & 46.3 & 54.1  & 15.3 & 34.4 & 51.1 & 58.5 \\
   JVTC \cite{li2020joint} (ECCV'20) &   20.3 & 45.4 & 58.4 & 64.3 & 19.0 & 42.1 & 53.4 & 58.9 \\
   GPR \cite{luogeneralizing} (ECCV'20) &  {20.4} & {43.7} & {56.1} & {61.9} & {24.3}& {51.7} &  {64.0} & {68.9} \\
   NRMT \cite{zhao2020unsupervised} (ECCV'20) &  19.8 & 43.7 & 56.5 & 62.2 & 20.6 & 45.2 & 57.8 & 63.3  \\
    \hline
    {UDAP~\cite{song2018unsupervised} w/ {DBSCAN} (modified)} & 20.3 & 45.7 & 58.6 & 64.5 & 22.8 & 49.3 & 61.7 & 67.2  \\
	Our SDA w/ {DBSCAN} & \textbf{23.2} & \textbf{49.5} & \textbf{62.2} & \textbf{67.7} & \textbf{25.6} & \textbf{54.4} & \textbf{66.4} & \textbf{71.3} \\
    Our SDA w/ {$k$-means}  &  {20.6} & {46.8} & {59.9} & {65.0} & {23.0} & {51.7} & {64.2} & {69.6} \\
	\hline
	MMT~\cite{ge2020mutual} w/ {$k$-means} (ICLR'20) & {22.9} & {49.2} & {63.1} & {68.8} & {23.3} & {50.1} & {63.9} & {69.8} \\
	Our SDA + MMT~\cite{ge2020mutual} w/ {$k$-means} & \textbf{29.0} & \textbf{57.0} & \textbf{69.5} & \textbf{74.1 } & \textbf{30.3} & \textbf{59.6} & \textbf{71.7} & \textbf{76.2} \\
	\hline
	\end{tabular}
	\end{center}
\end{table*}

\subsection{Comparison with State-of-the-arts}
\label{sec:sota}

We compare our proposed SDA framework with state-of-the-art methods on four domain adaptive re-ID tasks in Table \ref{tab:sota} and Table \ref{tab:sota1}.
Our method is plug-and-play with existing pseudo-label-based target domain encoders.
Note that the focus of our SDA is to generate informative training samples rather than pseudo label refinery as the previous methods. The translated images by our SDA are used as additional training samples to further improve the pseudo-label-based encoder.

\subsubsection{\textbf{ {Modified} UDAP~\cite{song2018unsupervised} as a baseline for the target-domain encoder}}
Following the label generation pipeline introduced by a clustering-based baseline method UDAP~\cite{song2018unsupervised},
we adopted DBSCAN \cite{ester1996density} to generate target-domain pseudo labels for our target-domain encoder.
 {Rather than using the sole triplet loss in the original UDAP~\cite{song2018unsupervised}, we modified it by using both classification loss and triplet loss as the training objective to achieve better baseline performance, dubbed as ``UDAP~\cite{song2018unsupervised} w/ DBSCAN (modified)'' in Table \ref{tab:sota}.  The detailed introduction of our modified training objective can be found at Eq. (\ref{eq:pic}).  Also note that, we denote our SDA model as ``Our SDA w/ {DBSCAN}'' instead of ``Our SDA + UDAP w/ {DBSCAN}'', since the SDA model is trained based on our modified version of UDAP instead of the original one.}
Our SDA outperforms the baseline UDAP~\cite{song2018unsupervised} by large margins, which indicates the effectiveness of our generated source-to-target training samples.

We also tested $k$-means to generate target-domain pseudo labels for our target-domain encoder on SDA with the optimal $k$ value following the state-of-the-art \cite{ge2020mutual}, \ie
$500$ for Duke$\to$Market, $700$ for Market$\to$Duke, $1500$ for Duke$\to$MSMT and Market$\to$MSMT.
The results are denoted as ``Our SDA w/ {$k$-means}''.
The training pipeline is the same as ``Our SDA w/ {DBSCAN}'', except for the clustering algorithm.
Our SDA is consistently effective without the need of setting $k$ to be close to the actual identity numbers.
As shown in Table \ref{tab:ks}, even with different $k$'s, our SDA stably improves the already strong {baselines}, which are trained with only the target-domain samples and clustering pseudo labels \cite{song2018unsupervised}.
 {Note that the value of $k$ can be considered as a hyper-parameter, and can be determined by searching for the optimal performance on the validation set. Selecting proper hyper-parameters for clustering-based methods would not limit their usage in practical use.}

\subsubsection{\textbf{MMT \cite{ge2020mutual} as a baseline for the target-domain encoder}}

Although MMT \cite{ge2020mutual} shows superior performances over ``Ours SDA w/ $k$-means'' by adopting dual networks with two times more parameters and computations for mutual training,
our SDA is well compatible with it and can be combined to achieve further improvements.
Specifically, the target-domain encoder is trained with source-to-target translated images and target-domain raw images under the training pipeline of MMT, where the source-to-target images are translated by SDA.
Both soft losses and hard losses in MMT are adopted for training.
The combination ``Our SDA w/ $k$-means+MMT \cite{ge2020mutual}" shows further $4.3\%$ and $5.3\%$ mAP gains on Duke$\to$Market and Market$\to$Duke.
 {Note that adopting existing domain translation-based methods (\eg SPGAN \cite{deng2018image}) on top of MMT \cite{ge2020mutual} even degrades the performance, since MMT \cite{ge2020mutual} itself is already very strong and can only be boosted by informative enough translated images.}


\begin{table}
\centering
\footnotesize
\begin{tabular}{P{0.6cm}|C{0.8cm}C{0.8cm}|C{1.5cm}C{1.5cm}}
	\hline
	\multicolumn{1}{c|}{\multirow{2}{*}{$k$ value}} & \multicolumn{2}{c|}{Baseline} & \multicolumn{2}{c}{Ours} \\
	\cline{2-5}
	\multicolumn{1}{c|}{} & mAP & top-1 & mAP & top-1 \\
	\hline \hline
   \centering 500 &  46.7 & 65.9 & 53.8 \textbf{(+7.1)}  & 70.6 \textbf{(+4.7)}  \\
   \centering 700 &   50.1 & 68.2  & 56.7 \textbf{(+6.6)} & 74.0 \textbf{(+5.8)} \\
  \centering  900 &   48.9 & 66.6  & 56.0 \textbf{(+7.1)} & 72.9 \textbf{(+6.3)}\\
	\hline
	\end{tabular}
\caption{Comparison with different values of $k$ in our SDA when adopting $k$-means on Market$\to$Duke.}\label{tab:ks}
\end{table}

\subsection{Comparison with Domain Translation-based Methods}
\label{sec:com_dt}
As previous translation-based methods \cite{deng2018image,wei2018person,deng2018similarity,chen2019instance} did not utilize pseudo labels and therefore cannot be directly compared with. We carefully design comparative experiments to verify the importance of the proposed online relation-consistency regularization term.

\subsubsection{\textbf{Our framework without pseudo labels}}
\label{sec:com_dt_1}
We evaluate our framework using a target-domain encoder without pseudo labels, which is a common strategy in previous translation-based methods \cite{wei2018person,deng2018image,liang2018m2m,chen2019instance}, \ie the target-domain encoder is trained with only source-to-target translated images and their source-domain identities.
As shown in Table \ref{tab:app_uda}, our method stably outperforms existing domain translation-based methods  {in terms of mAP} on both Duke$\to$Market and Market$\to$Duke adaptation tasks without generating pseudo labels in the target domain.
 {Specifically, we outperform state-of-the-art method CGAN-TM~\cite{tang2020cgan} by 3.7\% and 1.4\% mAP on Duke$\to$Market and Market$\to$Duke, respectively.  {Note that CGAN-TM~\cite{tang2020cgan} achieved slightly better top-1/5/10 performance than our method on Market$\to$Duke since they adopted a deeper DenseNet-121 as the backbone, compared to our plain ResNet-50 backbone.}}

 {In order to further verify the effectiveness of our introduced relation regularization term, we conduct an experiment by removing the $\mathcal{L}_\text{rc}$ from ``Our SDA w/o pseudo labels'', dubbed ``Our SDA w/o pseudo labels (w/o $\mathcal{L}_\text{rc}$)'' in Table \ref{tab:app_uda}. We observe significant performance drops when discarding such a relation regularization term from training.}

\begin{table*}[tb]
	\footnotesize
	\centering
	\caption{Comparison with domain translation-based UDA methods using target-domain encoder \textbf{without} pseudo labels. ``Our SDA w/o pseudo labels'' indicates training target-domain encoder with only source-to-target translated images and their ground-truth identities. }
	\label{tab:app_uda}
	\begin{center}
	\begin{tabular}{P{5cm}|C{1cm}C{1cm}C{1cm}C{1cm}|C{1cm}C{1cm}C{1cm}C{1cm}}
	\hline
	\multicolumn{1}{c|}{\multirow{2}{*}{Methods w/o Pseudo Labels}} & \multicolumn{4}{c|}{DukeMTMC-reID$\to$Market-1501} & \multicolumn{4}{c}{Market-1501$\to$DukeMTMC-reID} \\
	\cline{2-9}
	\multicolumn{1}{c|}{} & mAP & top-1 & top-5 & top-10 & mAP & top-1 & top-5 & top-10 \\
	\hline \hline
	PTGAN~\cite{wei2018person} (CVPR'18) &  - & 38.6 &-&-& - & 27.4& -& - \\
    SPGAN~\cite{deng2018image} (CVPR'18) & 22.8 & 51.5 & 70.1 & 76.8 & 22.3 & 41.1 & 56.6 & 63.0 \\
        CR-GAN~\cite{chen2019instance} (ICCV'19) & 29.6 & 59.6 & - & - & 30.0 & 52.2 & - & - \\
         {SBSGAN~\cite{huang2019sbsgan} (ICCV'19)} & 27.3 & 58.5 & - & - & 30.8 & 53.5 & - & - \\
         {CGAN-TM~\cite{tang2020cgan} (TIP'20) (DenseNet-121)} & 31.3 & 61.4 & 78.4 & \textbf{84.9} & 32.9 & \textbf{54.9} & \textbf{68.8} & \textbf{74.3} \\
        \hline
	Our SDA w/o pseudo labels & \textbf{35.0} & \textbf{64.5} & \textbf{79.5} & {84.6} & \textbf{34.3} & {53.1} & {67.1} & {72.4} \\ \hline
	 {Our SDA w/o pseudo labels (w/o $\mathcal{L}_\text{rc}$)} & 31.0 & 59.9 & 75.2 & 82.0 & 30.2 & 51.2 & 66.0 & 70.9 \\
	\hline
	\end{tabular}\\
	\end{center}
\end{table*}

\subsubsection{\textbf{Existing ID-based regularizations in our framework}}
\label{sec:com_dt_2}
Existing translation-based UDA methods \cite{deng2018image,wei2018person,deng2018similarity,chen2019instance} adopted identity-based losses with static targets to regularize the domain translation, including contrastive loss in SPGAN \cite{deng2018image}, classification loss in eSPGAN \cite{deng2018similarity} and CR-GAN \cite{chen2019instance}, and triplet loss in CGAN-TM \cite{tang2020cgan}. Generally, the previous losses only require the source-to-target translated images to be correctly classified after translation,  {but ignore the inter-sample relations and similarities in their original domain}.
Since our pseudo-label-based target-domain encoder shows much better baseline performance than theirs,
for fair comparison,
we replace the online relation regularization $\mathcal{L}_\text{rc}$ (Eq. (\ref{eq:sc})) in our framework with the previous methods' ID-based regularizations to demonstrate the effectiveness of our regularization.

The results are reported in Table \ref{tab:regul}.
We observe that replacing our proposed $\mathcal{L}_\text{rc}$ with previous regularization (denoted as ``$\mathcal{L}_\text{rc} \to$ ID-based contrastive loss \cite{deng2018image}'', ``$\mathcal{L}_\text{rc} \to$ ID-based classification loss  \cite{deng2018similarity,chen2019instance,zou2020joint}'',
``$\mathcal{L}_\text{rc} \to$ ID-based triplet loss \cite{tang2020cgan}'',  {``$\mathcal{L}_\text{rc} \to$ ID-based classification loss \& triplet loss''}) all lead to worse performances than our method, demonstrating the superiority of our stronger relation regularization term over the weaker regularizations in previous domain translation-based UDA methods.


\begin{table*}[t]
\footnotesize
	\centering
	\caption{Comparison between our online relation-consistency regularization $\mathcal{L}_\text{rc}$ and  ID-based regularizations in previous domain translation-based methods. $k$-means algorithm is adopted here to generate pseudo labels.}
	\label{tab:regul}
	\begin{center}
	\begin{tabular}{P{5.5cm}|C{1cm}C{1cm}C{1cm}C{1cm}|C{1cm}C{1cm}C{1cm}C{1cm}}
	\hline
	\multicolumn{1}{c|}{\multirow{2}{*}{Methods}} & \multicolumn{4}{c|}{DukeMTMC-reID$\to$Market-1501} & \multicolumn{4}{c}{Market-1501$\to$DukeMTMC-reID} \\
	\cline{2-9}
	\multicolumn{1}{c|}{} & mAP & top-1 & top-5 & top-10 & mAP & top-1 & top-5 & top-10 \\
	\hline \hline
    $\mathcal{L}_\text{rc} \to$ ID-based contrastive loss \cite{deng2018image} & 56.5 & 78.7 & 91.5 & 94.0 & 51.7 & 69.9 & 80.5 & 83.3 \\
    $\mathcal{L}_\text{rc} \to$ ID-based classification loss  \cite{deng2018similarity,chen2019instance,zou2020joint} & 63.4 & 84.9 & 92.7 & 95.1 & 53.8 & 70.9 & 81.9 & 85.6 \\
    $\mathcal{L}_\text{rc} \to$ ID-based triplet loss \cite{tang2020cgan} & 64.1 & 85.2 & 92.9 & 95.4 & 54.5 & 72.1 & 82.3 & 85.9 \\
     {$\mathcal{L}_\text{rc} \to$ ID-based classification loss \& triplet loss} & 64.5 & 85.0 & 93.1 & 95.6 & 54.7 & 72.6 & 82.4 & 85.9 \\
    \hline
	Our SDA w/ $\mathcal{L}_\text{rc}$ & \textbf{66.4} & \textbf{86.4} & \textbf{93.1} & \textbf{95.6} & \textbf{56.7} & \textbf{74.0} & \textbf{84.1} & \textbf{87.7} \\
	\hline
	\end{tabular}
	\end{center}
\end{table*}

\subsubsection{\textbf{Domain Translation Examples}}

\begin{figure*}[htb]
\centering
    \begin{tabular}{c@{\hspace{0.7mm}}c@{\hspace{0.7mm}}c@{\hspace{2mm}}@{\hspace{2mm}}c@{\hspace{0.7mm}}c@{\hspace{0.7mm}}c@{\hspace{2mm}}@{\hspace{2mm}}c@{\hspace{0.7mm}}c@{\hspace{0.7mm}}c@{\hspace{2mm}}@{\hspace{2mm}}c@{\hspace{0.7mm}}c@{\hspace{0.7mm}}c}

    \multicolumn{3}{c}{Source images} & \multicolumn{3}{c}{CycleGAN \cite{zhu2017unpaired} } & \multicolumn{3}{c}{SPGAN \cite{deng2018image}} & \multicolumn{3}{c}{Ours}  \\
        \includegraphics[scale=0.2]{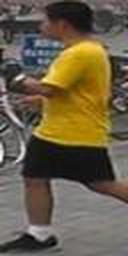} &
        \includegraphics[scale=0.2]{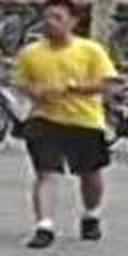} &
        \includegraphics[scale=0.2]{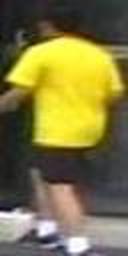} &

        \includegraphics[scale=0.2]{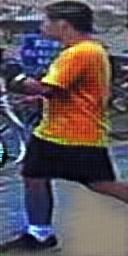} &
        \includegraphics[scale=0.2]{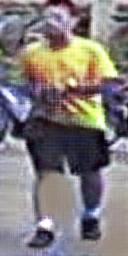} &
        \includegraphics[scale=0.2]{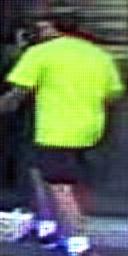} &

        \includegraphics[scale=0.2]{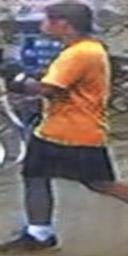} &
        \includegraphics[scale=0.2]{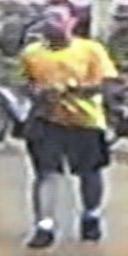} &
        \includegraphics[scale=0.2]{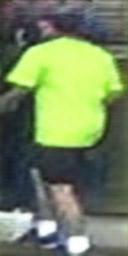} &

        \includegraphics[scale=0.2]{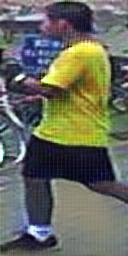} &
        \includegraphics[scale=0.2]{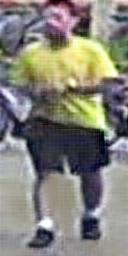} &
        \includegraphics[scale=0.2]{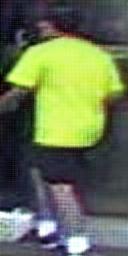} \\

        \includegraphics[scale=0.2]{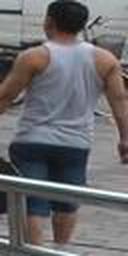} &
        \includegraphics[scale=0.2]{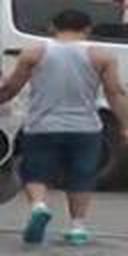} &
        \includegraphics[scale=0.2]{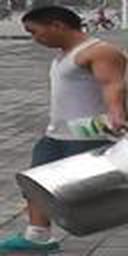} &

        \includegraphics[scale=0.2]{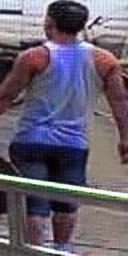} &
       \includegraphics[scale=0.2]{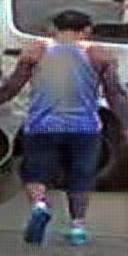} &
       \includegraphics[scale=0.2]{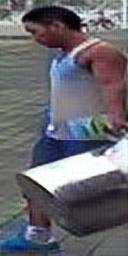} &

        \includegraphics[scale=0.2]{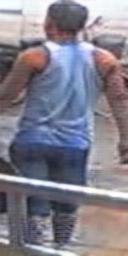} &
        \includegraphics[scale=0.2]{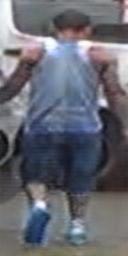} &
        \includegraphics[scale=0.2]{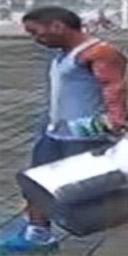} &

        \includegraphics[scale=0.2]{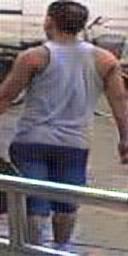} &
        \includegraphics[scale=0.2]{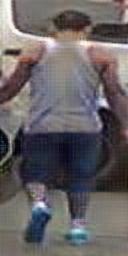} &
        \includegraphics[scale=0.2]{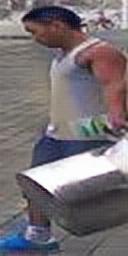} \\

        \includegraphics[scale=0.2]{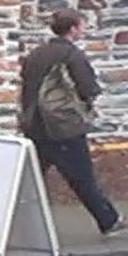} &
        \includegraphics[scale=0.2]{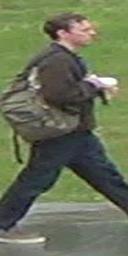} &
        \includegraphics[scale=0.2]{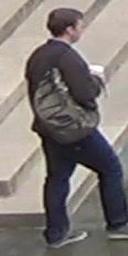} &

       \includegraphics[scale=0.2]{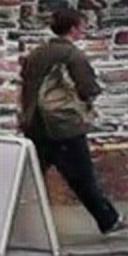} &
        \includegraphics[scale=0.2]{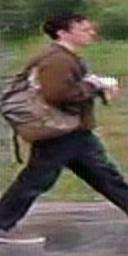} &
       \includegraphics[scale=0.2]{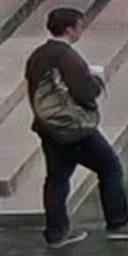} &

        \includegraphics[scale=0.2]{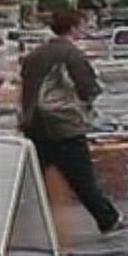} &
        \includegraphics[scale=0.2]{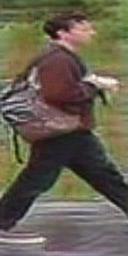} &
        \includegraphics[scale=0.2]{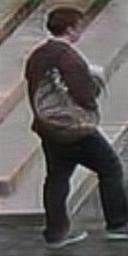} &

        \includegraphics[scale=0.2]{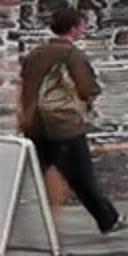} &
        \includegraphics[scale=0.2]{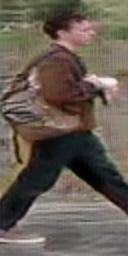} &
        \includegraphics[scale=0.2]{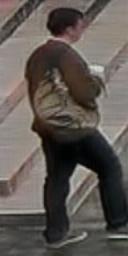} \\
    \end{tabular}
    \caption{Domain-translated examples of CycleGAN \cite{zhu2017unpaired}, SPGAN \cite{deng2018image} and our method.   {Note persons on each row are of the same identity.}
   Other translation-based methods \cite{wei2018person,deng2018similarity,chen2019instance} did not provide trained models or translated images, thus not be illustrated here.
   However, we have carefully discussed and compared them with their ID-based regularizations in Section \ref{sec:com_dt_1} \& \ref{sec:com_dt_2}.
    Best viewed in color.}
    \label{fig:gan}
\end{figure*}

Besides the example triplets as shown in Figure \ref{fig:fig1} (c),
we also illustrate several translated examples of CycleGAN \cite{zhu2017unpaired}, SPGAN \cite{deng2018image} and our method in Figure \ref{fig:gan}.
SPGAN adopts ID-based regularizations (\ie contrastive loss), showing inferior generation results than our method. ID-based regularizations are too weak to preserve inter-sample relations during translation. For instance, the man in the first row appears to be in different colors (\eg orange, yellow and green) within a tuple after translation by CycleGAN and SPGAN.

 {Besides the illustration of the generated images, we also evaluate the quality of translated samples by calculating the FID score \cite{heusel2017gans}. Specifically, we use FID metric to compute the similarity between the source-to-target translated dataset and the raw target-domain dataset. As demonstrated in Table \ref{tab:fid}, our method considerably outperforms SPGAN \cite{deng2018image} on the task of Market$\to$Duke. Note that the FID score can only evaluate the quality of generated images, but cannot evaluate the preserved inter-sample relations and affinities. One possible way to evaluate the preserved inter-sample relations of translated images is to use them as additional training samples for re-ID training and then evaluate the trained model by the re-ID metric, just as what we did in the above sections.}

\begin{table}
\centering
\footnotesize
\begin{tabular}{P{4cm}|C{1.5cm}C{1.5cm}}
	\hline
	\multicolumn{1}{c|}{} & SPGAN \cite{deng2018image} & Ours \\
	\cline{2-3}
	\hline
   \centering Market-1501$\to$DukeMTMC-reID &  47.6 & 77.6 \\
   \centering DukeMTMC-reID$\to$Market-1501 &   46.1 & 46.0 \\
	\hline
	\end{tabular}
\caption{ {Evaluation of generated images by domain translation-based methods for UDA person re-ID in terms of the FID score \cite{heusel2017gans}.}}\label{tab:fid}
\end{table}

\subsection{Ablation Studies}
\label{sec:abla}

\begin{table*}[t]
\footnotesize
	\centering
	\caption{Ablation studies for our proposed framework (``Our SDA w/ $k$-means'') on individual components.}
	\label{tab:ablation}
	\begin{center}
	\begin{tabular}{P{5.8cm}|C{1cm}C{1cm}C{1cm}C{1cm}|C{1cm}C{1cm}C{1cm}C{1cm}}
	\hline
	\multicolumn{1}{c|}{\multirow{2}{*}{Methods}} & \multicolumn{4}{c|}{DukeMTMC-reID$\to$Market-1501} & \multicolumn{4}{c}{Market-1501$\to$DukeMTMC-reID} \\
	\cline{2-9}
	\multicolumn{1}{c|}{} & mAP & top-1 & top-5 & top-10 & mAP & top-1 & top-5 & top-10 \\
	\hline \hline
    Source-domain pre-trained & 21.7 & 47.8 & 64.2 & 70.1 & 14.4 & 26.7 & 39.8 & 45.6 \\
    Baseline (only target-domain data + pseudo labels) & 59.0 & 80.7 & 90.5 & 93.4 & 50.1 & 68.2 & 79.2 & 82.7 \\
    Baseline + raw source-domain data & 53.2 & 77.5 & 88.9 & 92.1 & 50.8 & 69.1 & 80.1 & 83.1 \\
    \hline
    Ours w/o relation-consistency loss $\mathcal{L}_\text{rc}$ & 63.0 & 84.3 & 92.7 & 95.2 & 52.9 & 69.8 & 81.3 & 84.5 \\
    Ours w/o unified label system & 64.8 & 86.0 & 93.1 & 95.2 & 54.8 & 73.1 & 83.1 & 85.5 \\
    Ours w/o momentum encoder \cite{he2019momentum} & 65.3 & 86.1 & 93.1 & 95.3 & 55.3 & 72.5 & 82.8 & 85.5 \\
    \hline
	Ours (full) & \textbf{66.4} & \textbf{86.4} & \textbf{93.1} & \textbf{95.6} & \textbf{56.7} & \textbf{74.0} & \textbf{84.1} & \textbf{87.7} \\
	\hline
	\end{tabular}
	\end{center}
\end{table*}

We conduct ablation studies on Duke$\to$Market and Market$\to$Duke tasks to analyze the effectiveness of each component in our framework.
Detailed ablation experiments can be found in Table \ref{tab:ablation}.

\subsubsection{\textbf{Effectiveness of source-to-target translated images}}
\label{sec:compatible}

We treat the target-domain encoder $\mathcal{F}^t$ trained with only target-domain images and clustering-based pseudo labels as our \emph{baseline} model, which can be treated as a reproduction of UDAP \cite{song2018unsupervised}.
Our framework significantly outperforms the baseline model by properly exploiting valuable source-domain data (see ``Ours (full)'' vs. ``Baseline'' in Table \ref{tab:ablation}).

A na\"ive way to use source-domain images is to directly train on both domains' raw images, denoted as  ``Baseline+raw source-domain data''.
The performance is even worse than the baseline on Duke$\to$Market due to the large domain gaps, which indicates the necessity of properly leveraging different domains' images.

Our SDA can also be integrated and benefit state-of-the-art pseudo-label-based method \cite{ge2020mutual} (see ``Our SDA+MMT'' vs. ``MMT'' in Table \ref{tab:sota}).
The improvements demonstrate the effectiveness of source-to-target images translated by our structured domain translation method.

\subsubsection{\textbf{Alternative designs of online relation-consistency regularization}}
\label{sec:alter}

\begin{table}
\centering
\footnotesize
\begin{tabular}{P{3.4cm}|C{0.8cm}C{0.8cm}|C{0.8cm}C{0.8cm}}
	\hline
	\multicolumn{1}{c|}{\multirow{2}{*}{Regularization}} & \multicolumn{2}{c|}{Duke$\to$Market} & \multicolumn{2}{c}{Market$\to$Duke} \\
	\cline{2-5}
	\multicolumn{1}{c|}{} & mAP & top-1 & mAP & top-1 \\
	\hline \hline
    Prediction-consistency $\mathcal{L}_\text{pc}$ & 63.3 & 84.5 & 54.3 & 71.1 \\
    Batch-all relations $\mathcal{L}_\text{brc} \cite{tung2019similarity} $ & 64.6 & 86.0 & 54.6 & 71.5 \\
    \hline
	Our $\mathcal{L}_\text{rc}$ & \textbf{66.4} & \textbf{86.4}  & \textbf{56.7} & \textbf{74.0} \\
	\hline
	\end{tabular}
	\caption{Comparison with the optional relation-consistency regularizations in our SDA with $k$-means.}
\label{tab:options}
\end{table}

Our SDT applies regularizations on the softmax-triplet relations (Eq. (\ref{eq:sc})).
We further explore two alternative forms, prediction-consistency regularization $\mathcal{L}_\text{pc}$ and batch-all relation-consistency regularization $\mathcal{L}_\text{brc}$ (Table \ref{tab:options})
to verify the effectiveness of our proposed inter-sample relation constraint.

Specifically, the prediction-consistency regularization ensures that each individual image in the source domain should maintain the same ``soft'' class prediction after source-to-target translation.
The loss function is formulated as
\begin{align}
\mathcal{L}_\text{pc}(\mathcal{G}^{s\to t})=
\mathbb{E}_{x^s\sim \sX^s} [ - \mathcal{C}^s(\vf^s) \cdot \log (\mathcal{C}^t(\vf^{s \to t})) ]. \nonumber
\end{align}
As shown in Table \ref{tab:options},
$3.1\%$ and $2.4\%$ mAP drops are observed on the two tasks.
 {Here the prediction-consistency regularization $\mathcal{L}_\text{pc}$ is different from conventional ID-based classification regularization \cite{deng2018similarity,chen2019instance,zou2020joint}, which can be formulated as $\mathbb{E}_{x^s\sim \sX^s} [ - y^s \cdot \log (\mathcal{C}^t(\vf^{s \to t})) ]$ with ``one-hot'' ground-truth label $y^s$.   Briefly, $\mathcal{L}_\text{pc}$ is supervised by ``soft'' $\mathcal{C}^s(\vf^s)$ while ID-based classification regularization is supervised by ``hard'' $y^s$. $\mathcal{C}^s(\vf^s)$ is predicted by the source-domain encoder, which has captured the original inter-sample relations in the latent space. Note in this work, we denote the detailed data distributions and structured relations among data points, rather than the simple positive and negative identity relations, as the inter-sample relations.  $\mathcal{L}_\text{pc}$ can better preserve the inter-sample relations, while the ID-based classification regularization can only ensure that images belonging to different identities are well separated.}

Our $\mathcal{L}_\text{rc}$ (Eq. (\ref{eq:sc})) aims at preserving relations within \textit{hardest} triplets,
while the alternative \textit{batch-all} relation-consistency loss  $\mathcal{L}_\text{brc}$ tries to preserve all available relations within batches, no matter whether they are easy or hard.
 {The batch-all regularization term follows the similarity-preserving loss in \cite{tung2019similarity}, which preserves inter-sample affinities in knowledge distillation.}
We model the batch-all inter-sample relations by measuring
the similarities
\begin{align}
\mathcal{R}(x^s;\mathcal{F}^s)= {\text{softmax}}[\langle \vf^s, \vf_1^s \rangle,\cdots,\langle \vf^s, \vf_k^s \rangle]. \nonumber
\end{align}
The term consists of pairwise dot products between each sample $x^s$ and all other ones in the same batch.
The similarity vector is normalized by a softmax function and a soft cross-entropy loss is adopted to regularize all the relations after translation,
\begin{align}
\mathcal{L}_\text{brc}(\mathcal{G}^{s\to t})=\mathbb{E}_{x^s\sim \sX^s} [ - \mathcal{R}(x^s; \mathcal{F}^s) \cdot \log \mathcal{R}(x^s; \mathcal{G}^{s \to t}, \mathcal{F}^t) ]. \nonumber
\end{align}
1.8\% mAP and 2.1\% mAP drops can be observed on the two tasks.
The reason might be that
batch-all relations contain many easy cases that cannot provide effective supervisions for training.

To show the necessity of adopting relation regularization during translation, we also tested totally removing $\mathcal{L}_\text{rc}$ from our framework, dubbed ``Ours w/o relation-consistency loss $\mathcal{L}_\text{rc}$'' in Table \ref{tab:ablation}.
Significant mAP decreases of $3.4\%$ and $3.8\%$ are observed on Duke$\to$Market and Market$\to$Duke tasks.
 {We observe that,  replacing our relation-consistency loss $\mathcal{L}_\text{rc}$ with either conventional ID-based constraints (Table \ref{tab:regul}) or alternative designs of relation-consistency regularization (Table \ref{tab:options}) would achieve only slight improvements over ``Ours w/o relation-consistency loss $\mathcal{L}_\text{rc}$''.  This is because the pseudo-label-based baseline is already strong enough to achieve satisfactory performance on the target domain. Further improvements are quite
challenging and require informative enough translated images as additional training samples. The comparison results well validate the effectiveness of our proposed relation-consistency regularization.}

\subsubsection{\textbf{Effectiveness of training with the unified label set}}
We observe that the target-domain encoder also benefits from the unified label set by training the classifier on all the $p^s+\hat{p}^t$ classes across the two domains.
To show it, we design an experiment with separate classifiers for source-to-target translated images and target-domain images,
\ie ${{\cal C}^t}:\vf\rightarrow\{1,\cdots, p^s+\hat{p}^t \}$ is split into $\mathcal{C}^{s\to t}:\vf^{s\to t} \rightarrow\{1,\cdots, p^s\}$ and $\mathcal{C}^t:\vf^t \rightarrow\{1,\cdots, \hat{p}^t \}$.
We report the performance in Table \ref{tab:ablation} as ``Ours w/o unified label system''.
1.6\% and 1.9\% mAP drops are observed on the two tasks, which indicate the effectiveness of modeling the relations between two domains.

\subsubsection{\textbf{Further benefits from the momentum encoder $\mathcal{F}^t_*$}}
As described in Section \ref{sec:imp}, we utilize a momentum encoder \cite{he2019momentum} for more stable training and better performance.
To verify that the main contribution is not from the momentum encoder,
we perform an experiment by removing  $\mathcal{F}^t_*$ while keeping all other components unchanged.
The experimental results are denoted as ``Ours w/o momentum encoder \cite{he2019momentum}'' in Table \ref{tab:ablation}.
We observe slight drops of 1.1\% and 1.4\% mAP on two tasks.

\subsubsection{\textbf{Performance w/o joint training of domain-translation network and target-domain encoder}}
\label{sec:woj}

\begin{table*}[htb]
\footnotesize
	\centering
	\caption{Comparison with domain translation-based UDA methods with pseudo labels (via $k$-means clustering) but {\bf without} jointly training the domain-translation network and target-domain encoder. Source-to-target images translated by SPGAN \cite{deng2018image} were provided by the authors and can be directly used for training the pseudo-label-based encoder.}
	\label{tab:app_abla}
	\begin{center}
	\begin{tabular}{P{5.8cm}|C{1cm}C{1cm}C{1cm}C{1cm}|C{1cm}C{1cm}C{1cm}C{1cm}}
	\hline
	\multicolumn{1}{c|}{\multirow{2}{*}{Methods w/o Joint Training}} & \multicolumn{4}{c|}{DukeMTMC-reID$\to$Market-1501} & \multicolumn{4}{c}{Market-1501$\to$DukeMTMC-reID} \\
	\cline{2-9}
	\multicolumn{1}{c|}{} & mAP & top-1 & top-5 & top-10 & mAP & top-1 & top-5 & top-10 \\
	\hline \hline
    Baseline (target-domain data + pseudo labels) & 59.0 & 80.7 & 90.5 & 93.4 & 50.1 & 68.2 & 79.2 & 82.7 \\
    \hline
    Base. + source-to-target data by CycleGAN & 56.0 & 79.6 & 90.6 & 93.9 & 51.2 & 69.5 & 80.4 & 83.5 \\
    Base. + source-to-target data by SPGAN \cite{deng2018image} & 53.4 & 78.6 & 90.3 & 93.1 & 48.8 & 66.2 & 78.5 & 83.0 \\
    \hline
    Base. + source-to-target data by our SDT & \textbf{61.3} & \textbf{83.3} & \textbf{91.8} & \textbf{94.9} & \textbf{54.3} & \textbf{71.6} & \textbf{82.0} & \textbf{85.6} \\
	\hline
	\end{tabular}
	\end{center}
\end{table*}

The domain-translation network and target-domain encoder in our framework promote each other via joint training. However, a simpler training scheme would be to first train a source-to-target translation network with the proposed regularization and to translate all source-domain images to the target domain.
A pseudo-label-based target-domain encoder is then trained  with such fixed source-to-target translated images and target-domain images.
Note that in this scheme, a target-domain encoder pre-trained with only unlabeled data and clustering labels is used for regularizing the SDT training.


We evaluate both our framework and existing translation-based methods \cite{zhu2017unpaired,deng2018image} when adopting such a separate training strategy and $k$-means clustering for pseudo label generation.
The results in Table \ref{tab:app_abla} show that the informative training samples generated by our proposed structured domain-translation (SDT) network could effectively improve the already strong baseline even without our joint training scheme, while the source-to-target images generated by CycleGAN and SPGAN might even worsen the performance because their generated images might not well capture the distributions of target-domain data and maintain their original inter-sample relations.

\subsubsection{\textbf{ {Hyper-parameter analysis}}}
\begin{figure}
    \centering
    \includegraphics[width=1.0\linewidth]{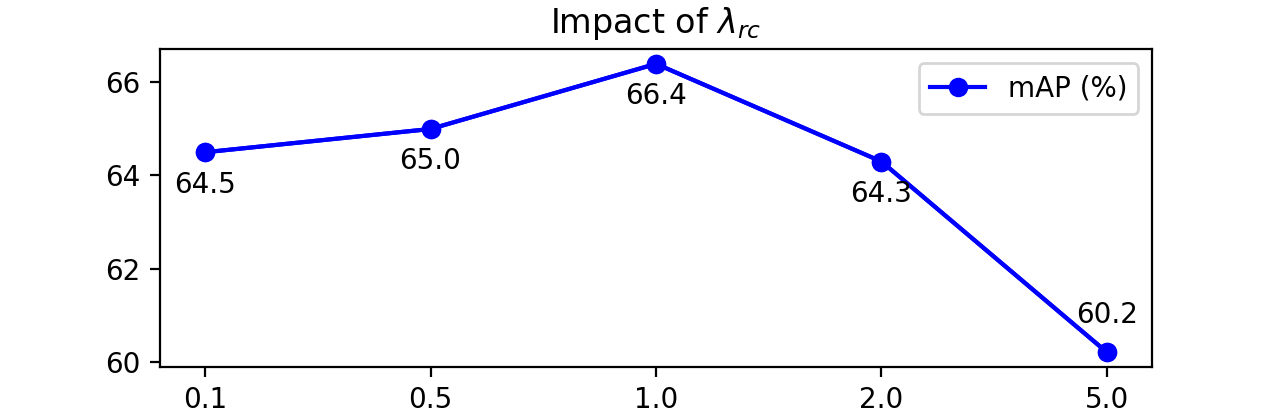}
    \caption{Hyper-parameter analysis on the loss weight $\lambda_\text{rc}$ for our proposed framework (``Our SDA w/ $k$-means'') on the task of Duke$\to$Market.}
    \label{fig:rc}
\end{figure}

 {
There are five hyper-parameters in our training scheme (Section \ref{sec:imp}), including the loss weights $\lambda_\text{rc}$, $\lambda_\text{adv}$, $\lambda_\text{cyc}$, $\lambda_\text{apr}$ and the triplet margin $m$. Note that the values of $\lambda_\text{adv}$, $\lambda_\text{cyc}$ and $\lambda_\text{apr}$ are directly inherited from CycleGAN \cite{zhu2017unpaired}, and the setting of $m=0.3$ is widely acknowledged in re-ID-related tasks \cite{luo2019bag,ge2020mutual}. We therefore conduct the experiments of hyper-parameter analysis on $\lambda_\text{rc}$, which is also the most important factor in our introduced structured domain-translation network. As shown in Figure \ref{fig:rc}, our proposed model is robust when the value of $\lambda_\text{rc}$ varies from $0.1$ to $2.0$.}

\subsection{Application of SDA in VisDA-2020 Challenge}

\begin{figure}[t]
\centering
\includegraphics[width=1.0\linewidth]{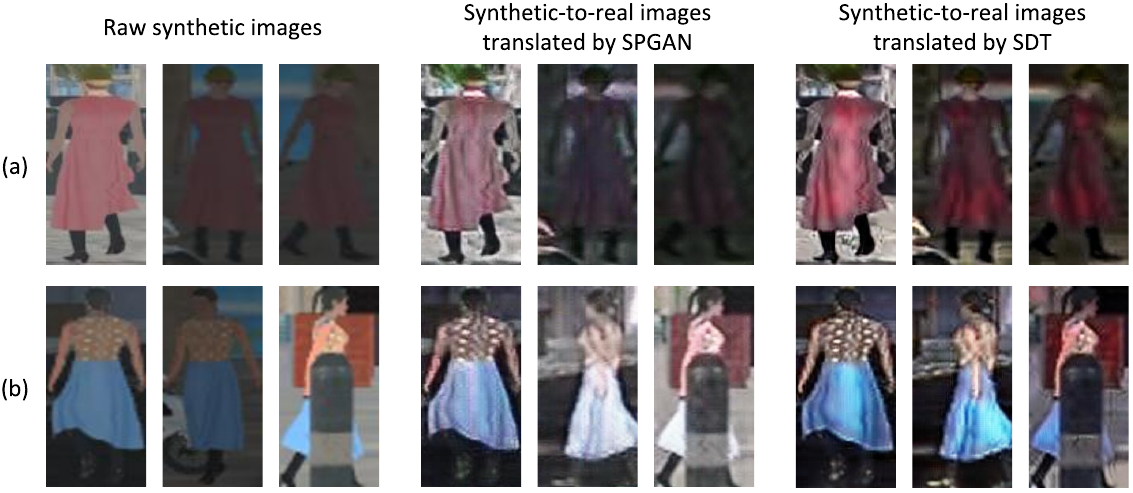}
\caption{Source-to-target translated images in VisDA-2020.   {Note persons on each row are of the same identity.}
Due to the large illumination variations,
(a) the woman in red appears in both red and black, and (b) another woman in blue appears in both blue and white after being translated by SPGAN \cite{deng2018image}.
In contrast, thanks to the online relation-consistency regularization, our SDT could better maintain a consistent appearance for person images of the same identity in both two cases.
}
\label{fig:sda}
\end{figure}

VisDA-2020 Challenge introduces a synthetic$\to$real adaptation task, where the labeled synthetic images and the unlabeled real-world images are provided.
 {The final performances on the test set of the real-world domain are used for ranking the teams in the challenge. Note that the model's performance on the test set is evaluated on the online server and the ground-truth labels cannot be accessed.  We therefore tune our models on the validation set during the challenge, and also report the performance on the validation set for comparison in this paper.}


We propose to adopt the structured domain-translation (SDT) network to translate the synthetic images to have the real-world style.
Specifically, we adopt the ``w/o joint training'' version of SDA as mentioned in Section \ref{sec:woj} with three main training steps:
1) Pre-training the source-domain encoder with only labeled data;
2) Pre-training the target-domain encoder with only unlabeled data and clustering labels;
3) Training the SDT network with the online relation-consistency regularization provided by the two encoders on-the-fly via Eq. (\ref{eq:sdt}). The main reasons why we did not use the ``joint training'' version of SDA for the competition is that increasing model sizes and numbers of parameters can result in more gains on the dataset, which is important in the competition. ``w/o joint training'' can save GPU memory and allow training much larger deep models.

After being trained, the SDT network can be adopted to generate informative training samples with ground-truth identities by translating all the source-domain images to the target domain.
In order to verify the effectiveness of our SDT, we adopt a target-domain encoder with a ResNet50-IBN backbone.
The encoder is trained with only synthetic-to-real translated images and their ground-truth identities, and then tested on the validation set of the target domain.
As shown in Table \ref{tab:sda}, training with ``synthetic-to-real images by SDT'' could achieve much more performance gain than training with ``synthetic-to-real images by SPGAN''.
Besides the quantitative comparison, we also visualize the translation results in Figure \ref{fig:sda}, where our SDT could better preserve the inter-sample relations than SPGAN \cite{deng2018image}.

\begin{figure}
    \centering
    \includegraphics[width=1.0\linewidth]{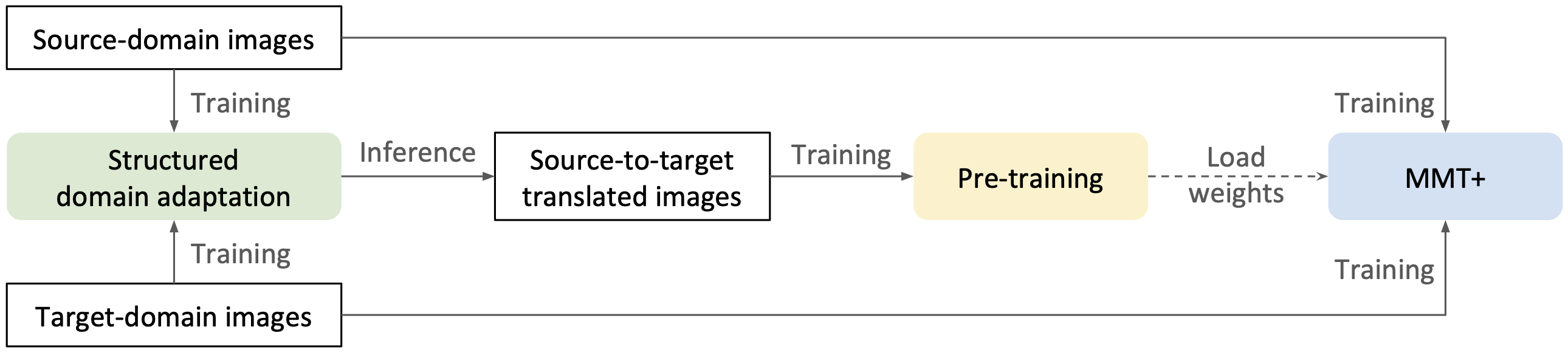}
    \caption{{The flow chart of our solution in VisDA-2020 Challenge, which consists of three steps: structured domain adaptation (SDA), pre-training with source-to-target translated images and fine-tuning on the target domain with the improved MMT framework.}}
    \label{fig:flow}
\end{figure}

In our solution to VisDA-2020 Challenge, multi-model ensemble and an improved MMT \cite{ge2020improved,ge2020mutual} are further introduced to fine-tune the network with both synthetic$\to$real images translated by SDT and raw unlabeled real-world images, achieving superior performance, as illustrated in Figure \ref{fig:flow}. The gain of our proposed SDT over the new baseline model and new adaption task demonstrates its generalization capability.
Our final solution ranks second out of 153 teams.

\begin{table}[tbh]
\footnotesize
	\caption{Ablation study on the effectiveness of synthetic-to-real images translated by SDT in VisDA-2020. The experiments adopt the backbone of ResNet50-IBN \cite{pan2018two} and post-processing techniques (\eg camera bias subtraction \cite{zhu2020voc}, re-ranking \cite{zhong2017re}) are employed. The results are evaluated on the target\_val set and only the top-100 matches are considered for mAP. Note that this result is not the final result in the challenge.}
	\label{tab:sda}
	\begin{center}
	\begin{tabular}{P{5cm}|C{1cm}C{1cm}}
	\hline
	Images for training & mAP & top-1 \\
	\hline \hline
	Raw synthetic images & 61.0 & 71.6 \\
	Synthetic-to-real images by SPGAN \cite{deng2018image} & 68.2 & 75.1 \\
	\textbf{Synthetic-to-real images by SDT} & \textbf{71.2} &	\textbf{79.3} \\
	\hline
	\end{tabular}
	\end{center}
\end{table}

\section{Discussion and Conclusion}

In this work, we propose an end-to-end structured domain adaptation framework with a novel online relation-consistency regularization term to tackle the unsupervised domain adaptation (UDA) problem for person re-ID. The structuredly translated images in our method are shown to be informative samples for improving the training of pseudo-label-based encoder. The joint optimization scheme of domain-translation network and re-ID encoder is effective, however, it still has difficulty on handling industrial-scale datasets. Further improvements are called for. Beyond the person re-ID, our proposed inter-sample relation-consistency regularization may benefit other related UDA tasks.




%

%

%
%

\ifCLASSOPTIONcaptionsoff
  \newpage
\fi



%
%
%

\bibliographystyle{IEEEtran}
\bibliography{TNNLS-2021-P-16392.R1}

%


\begin{IEEEbiography}[{\includegraphics[width=1in,height=1.25in,clip,keepaspectratio]{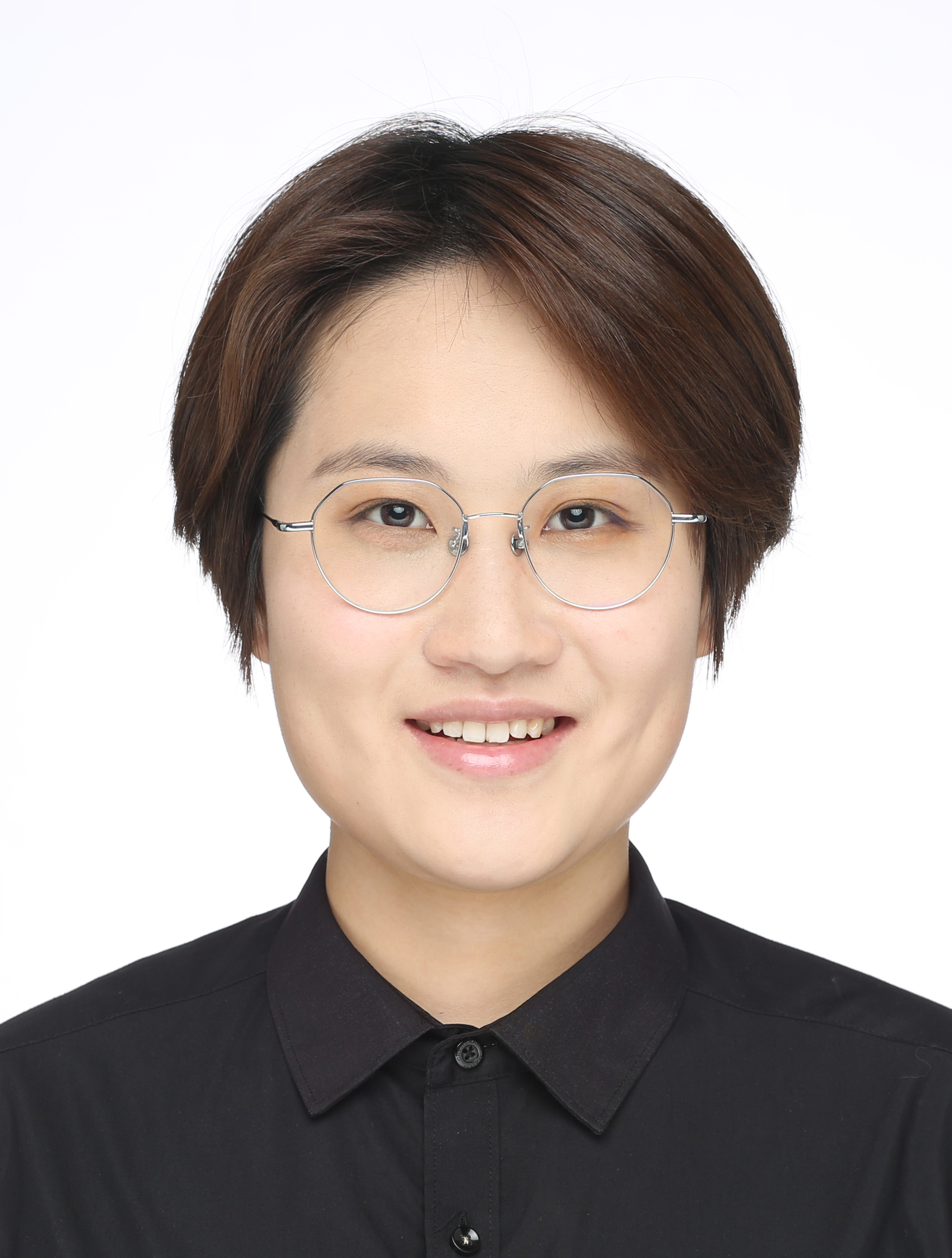}}]{Yixiao Ge}
received her B.S. degree from Huazhong University of Science and Technology in 2017, and the Ph.D. degree from Multimedia Laboratory, The Chinese University of Hong Kong in 2021.
She is currently with ARC Lab of Tencent PCG.
Her research interests include computer vision and deep learning, with focus on representation learning.
\end{IEEEbiography}

\begin{IEEEbiography}[{\includegraphics[width=1in,height=1.25in,clip,keepaspectratio]{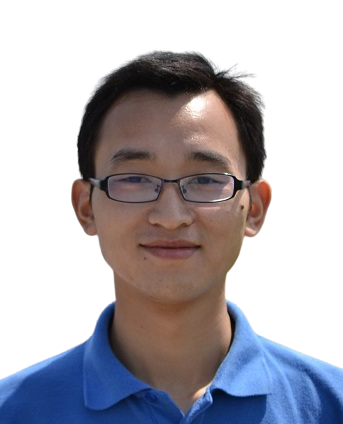}}]{Feng Zhu}
received his B.S. degree in electronic information engineering and Ph. D. degree in information and communication engineering in 2011 and 2017, respectively, both from the University of Science and Technology of China.
He is currently with Multimedia Laboratory, The Chinese University of Hong Kong.
His research interests include computer vision, face analysis, and machine learning.
\end{IEEEbiography}

\begin{IEEEbiography}[{\includegraphics[width=1in,height=1.25in,clip,keepaspectratio]{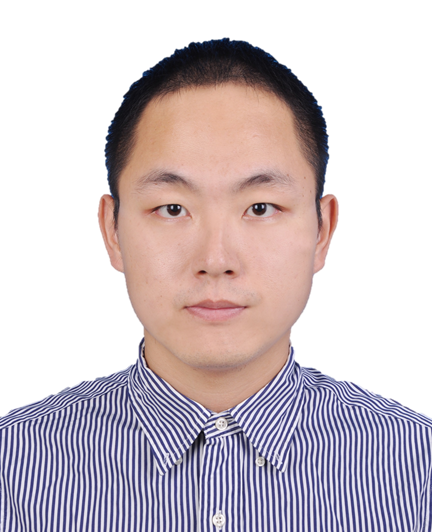}}]{Dapeng Chen}
received his B.S. degree in biomedical engineering and Ph.D. degree in control science and engineering from Xi'an Jiaotong University, China in 2010 and 2016, respectively.
He is currently with Multimedia Laboratory, The Chinese University of Hong Kong.
His research interests include computer vision, machine learning and large-scale retrievel/clustering.
\end{IEEEbiography}

\begin{IEEEbiography}[{\includegraphics[width=1in,height=1.25in,clip,keepaspectratio]{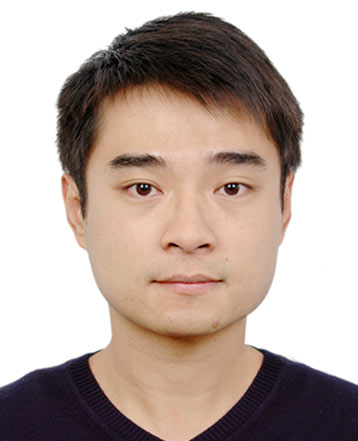}}]{Rui Zhao}
received his B.S. degree in electronic engineering and information science from University of Science and Technology of China, and Ph.D. degree in electronic engineering from the Chinese University of Hong Kong, China, in 2010 and 2015, respectively.
He is currently with Multimedia Laboratory, The Chinese University of Hong Kong.
His research interests include computer vision and machine learning, with focus on face analysis, deep learning, and visual surveillance.
\end{IEEEbiography}



\begin{IEEEbiography}[{\includegraphics[width=1in,height=1.25in,clip,keepaspectratio]{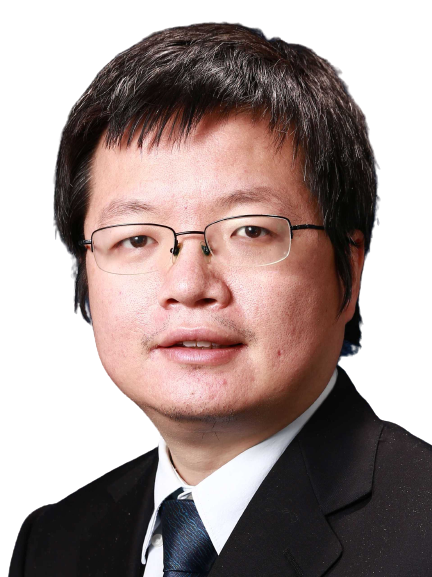}}]{Xiaogang Wang}
received the B.S. degree from the University of Science and Technology of China in 2001, the M.S. degree from The Chinese University of Hong Kong in 2003, and the Ph.D. degree from the Computer Science and Artificial Intelligence Laboratory, Massachusetts Institute of Technology in 2009. He is currently a professor in Multimedia Laboratory at The Chinese University of Hong Kong. His research interests include computer vision and machine learning.
\end{IEEEbiography}

\begin{IEEEbiography}[{\includegraphics[width=1in,height=1.25in,clip,keepaspectratio]{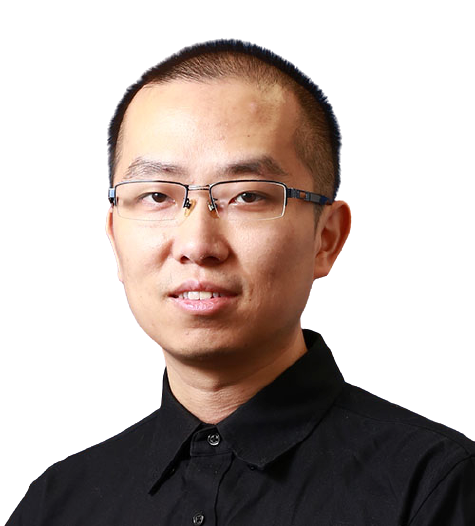}}]{Hongsheng Li}
received his B.S. degree in automation from the East China University of Science and Technology, and M.S. and Ph.D. degrees in computer science from Lehigh University, Pennsylvania, in 2006, 2010, and 2012, respectively. He is currently an assistant professor in Multimedia Laboratory at the Chinese University of Hong Kong. His research interests include computer vision, medical image analysis, and machine learning.
\end{IEEEbiography}




\end{document}